\documentclass[11pt]{article}
\usepackage{graphicx}
\usepackage{amssymb}
\usepackage{amscd}
\usepackage{mathrsfs}
\usepackage{longtable,lscape}
\usepackage{amsthm}
\usepackage{amsfonts}
\usepackage{amsmath}
\usepackage{bbm}
\usepackage{float}
\usepackage{url}

\newcommand{\captionfonts}{\footnotesize}
\makeatletter  
\long\def\@makecaption#1#2{%
  \vskip\abovecaptionskip
  \sbox\@tempboxa{{\captionfonts #1: #2}}%
  \ifdim \wd\@tempboxa >\hsize
    {\captionfonts #1: #2\par}
  \else
    \hbox to\hsize{\hfil\box\@tempboxa\hfil}%
  \fi
  \vskip\belowcaptionskip}
\makeatother 
\begin{document}
\title{Generalizing Prototype Theory: A Formal Quantum Framework}
\author{Diederik Aerts$^1$, Jan Broekaert$^1$, Liane Gabora$^{2}$, Sandro Sozzo$^{3}$ \vspace{0.5 cm} \\ 
        \normalsize\itshape
        $^1$ Center Leo Apostel for Interdisciplinary Studies, 
         Brussels Free University \\ 
        \normalsize\itshape
         Krijgskundestraat 33, 1160 Brussels, Belgium \\
        \normalsize
        E-Mails: \url{diraerts@vub.ac.be,jbroekae@vub.ac.be}
          \vspace{0.5 cm} \\ 
        \normalsize\itshape
        $^2$ Department of Psychology, University of British Columbia, Okanagan Campus \\
        \normalsize\itshape
        3333 University Way, Kelowna, BC Canada V1V 1V7
        \\
        \normalsize
        E-Mail: \url{liane.gabora@ubc.ca}
          \vspace{0.5 cm} \\ 
        \normalsize\itshape
        $^3$ School of Management and IQSCS, University of Leicester \\ 
        \normalsize\itshape
         University Road, LE1 7RH Leicester, United Kingdom \\
        \normalsize
        E-Mail: \url{ss831@le.ac.uk} \\
              }
\date{}
\maketitle
\begin{abstract}
\noindent
Theories of natural language and concepts have been unable to model the flexibility, creativity, context-dependence, and emergence, exhibited by words, concepts and their combinations. The mathematical formalism of quantum theory has instead been successful in capturing these phenomena such as graded membership, situational meaning, composition of categories, and also more complex decision making situations, which cannot be modeled in traditional probabilistic approaches. We show how a formal quantum approach to concepts and their combinations can provide a powerful extension of prototype theory. We explain how prototypes can interfere in conceptual combinations as a consequence of their contextual interactions, and provide an illustration of this using an intuitive wave-like diagram. This quantum-conceptual approach gives new life to original prototype theory, without however making it a privileged concept theory, as we explain at the end of our paper.
\end{abstract}

\section{Introduction}\label{intro}
Theories of concepts struggle to capture the creative flexibility with which concepts are used in natural language, and combined into larger complexes with emergent meaning, as well as the context-dependent manner in which concepts are understood  \cite{g1989}. In this paper, we present some recent advances in our quantum approach to concepts. More specifically, we follow the general lines illustrated in \cite{ga2002,ag2005a,ag2005b,gra2008}, and generalize the quantum-theoretic model elaborated in \cite{a2009b,ags2013}.

According to the `classical', or `rule-based' view of concepts, which can be traced back to Aristotle, all instances of a concept share a common set of necessary and sufficient defining properties. Wittgenstein pointed out that: (i) in some cases it is not possible to give a set of characteristics or rules defining a concept; (ii) it is often unclear whether an object is a member of a particular category; (iii) conceptual membership of an instance strongly depends on the context.

A major blow to the classical view came from Rosch's work on color.  This work showed that  colors do not have any particular criterial attributes or definite boundaries, and instances differ with respect to how typical they are of a concept \cite{r1973,r1978,r1983}. This led to formulation of `prototype theory', according to which concepts are organized around family resemblances, and consist of characteristic, rather than defining,  features. These features are weighted in the definition of the `prototype'. Rosch showed that subjects rate conceptual membership as `graded', with degree of membership of an instance corresponding to conceptual distance from the prototype. Moreover, the prototype appears to be particularly resistant to forgetting. Prototype theory also has the strength that it can be mathematically formulated and empirically tested. By calculating the similarity between the prototype of a concept and a possible instance of it, across all salient features, one arrives at a measure of the `conceptual distance' between the instance and the prototype. Another means of calculating conceptual distance comes out of `exemplar theory' \cite{n1988,n1992}, according to which a concept is represented by, not a set of defining or characteristic features, but a set of salient `instances' of it stored in memory. Exemplar theory has met with considerable success at predicting empirical results. Moreover, there is evidence of preservation of specific training exemplars in memory. Classical, prototype, and exemplar theories are sometimes referred to as `similarity based' approaches, because they assume that categorization relies on data-driven statistical evidence. They have been contrasted with `explanation based' approaches, according to which categorization relies on a rich body of knowledge about the world. For example, according to `theory theory' concepts take the form of `mini-theories' \cite{mm1985} or schemata \cite{rn1988}, in which the causal relationships among properties are identified.

Although these theories do well at modeling empirical data when only one concept is concerned, they perform poorly at modeling combinations of two concepts. As a consequence, cognitive psychologists are still looking for a satisfactory and generally accepted model of how concepts combine.

The inadequacy of fuzzy set models of conceptual conjunctions \cite{z1982} to resolve the `Pet-Fish problem' identified by Osherson and Smith \cite{os1981} highlighted the severity of the combination problem. People rate the item {\it Guppy} as a very typical example of the conjunction {\it Pet-Fish}, without rating {\it Guppy} as a typical example neither of {\it Pet} nor of {\it Fish} (`Guppy effect') \cite{os1981,os1982}. Studies by Hampton on concept conjunctions \cite{h1988a}, disjunctions \cite{h1988b} and negations \cite{h1997} confirmed that traditional fuzzy set and Boolean logical rules are violated whenever people combine concepts, as one usually finds `overextension' and `underextension' in the membership weights of items with respect to concepts and their combinations.  It has been shown that people estimate a sentence like ``$x$ is tall and $x$ is not tall'' as true, in particular when $x$ is a `borderline case' (`borderline contradictions')
(\cite{bovw1999,ap2011}), again violating the rules of set-theoretic Boolean logic. The seriousness of the combination problem was pointed out by various scholars \cite{k1992,f1994,kp1995,r1995,os1997}. More recently, other theories of concepts have been developed, such as `Costello and Keane's constraint theory' \cite{ck2000}, `Dantzig, Raffone, and Hommel's  connectionist CONCAT model of concepts' \cite{drh2011}, `Thagard and Stewart's emergent binding model' \cite{ts2011}, and `Gagne and Spalding's morphological approach' \cite{gs2009}. However none of these theories has a strong track record of modeling the emergence and non-compositionality of concept combinations. 

The approach to concepts presented in this paper grew out  earlier work on the application to concept theory on the axiomatic and operational foundations of quantum theory and quantum probability \cite{a1986,p1989,a1999}. A major theoretical insight was to shift the perspective from viewing a concept as a `container'  to viewing it as `an entity in a specific state that is changing under the influence of a context'  \cite{ga2002}. This allowed us to provide a solution to the Guppy effect and to successfully represent the data collected on {\it Pet}, {\it Fish} and {\it Pet-Fish} by using the mathematical formalism of quantum theory \cite{ag2005a,ag2005b}. Then, we proved that none of the above experiments in concept theory can be represented in a single probability space satisfying the axioms of Kolmogorov \cite{k1933}. We developed a general quantum framework to represent conjunctions, disjunctions and negations of two concepts, which has been successfully tested several times \cite{a2009b,a2009a,s2014a,s2014b,asv2015}, and we put forward an explanatory hypothesis for the observed deviations from traditional logical and probabilistic structures and for the occurrence of quantum effects in cognition \cite{IQSA2}. We recently identified a strong and systematic non-classical phenomenon effect, which is deeper than the ones typically detected in concept combinations and directly connected with the mechanisms of concept formation \cite{PhilTransA2015}. This work is part of a growing domain of cognitive psychology that uses the mathematical formalism of quantum theory and quantum structures to model empirical situations where the application of traditional probabilistic approaches is problematical (probability judgments errors, decision-making errors, violations of expected utility theory, etc.) \cite{ags2013,aa1995,aabg2000,as2011,bb2012,abgs2013,hk2013,pb2013,as2014,ast2014,pnas,plosone,IQSA1}.

This paper outlines recent progress in the development of a quantum-theoretic framework for concepts and their dynamics. Section \ref{contextualprototype} explains how the `SCoP formalism' can be interpreted as a `contextual and interfering prototype theory that is a generalization of standard prototype theory' in which prototypes are not fixed, but change under the influence of a context, and interfere as a consequence of their contextual interactions (see also \cite{gra2008,ags2013}). Section \ref{quantummodel} presents an amended explanatory version of the quantum-mechanical model  in complex Hilbert space worked out in \cite{a2009b,ags2013} for the typicality of items with respect to the concepts {\it Fruits} and {\it Vegetables}, and their disjunction {\it Fruits or Vegetables}. This improved quantum model illustrates how the prototype of {\it Fruits} ({\it Vegetables}) changes under the influence of the context  {\it Vegetables} ({\it Fruits}) in the combination {\it Fruits or Vegetables}. The latter combination is represented using the quantum-mathematical notion of linear superposition in a complex Hilbert space, which entails the genuine  quantum effect of `interference'. Hence, our model shows that the prototypes of {\it Fruits} and {\it Vegetables} interfere in the disjunction {\it Fruits or Vegetables}. Sections \ref{contextualprototype} and \ref{quantummodel} also justify the fact that our quantum-theoretic framework for concepts can be considered as a `contextual and interfering generalization of original prototype theory'. The presence of linear superposition and interference could suggest that concepts combine and interact like waves do. In Section \ref{geometric} we develop this intuition in detail and propose an intuitive wave-like illustration of the disjunction {\it Fruits or Vegetables}. Finally, Section \ref{conclusions} discusses connections between the quantum-theoretic approach to concepts presented here, and other theories of concepts. Although this approach can be interpreted as a  specific generalization of prototype theory, it is compatible with insights from other theories of concepts.

We stress that our investigation does not deal with the elaboration of a `specific typicality model' that represents a given set of data on the concepts {\it Fruits}, {\it Vegetables}, and their disjunction {\it Fruits or Vegetables}. We inquire into the mathematical formalism of quantum theory as a general, unitary and coherent formalism to model natural concepts. Our quantum-theoretic model in Section \ref{quantummodel} has been derived from this general quantum theory, hence it satisfies specific technical and general epistemological constraints of quantum theory. As such, it does not apply to any arbitrary set of experimental data. Our formalism exactly applies to those data that exhibit a peculiar deviation from classical set-theoretic modeling; such deviations are taken in our framework as indicative of interference and emergence. Data collected on combinations of two concepts systematically exhibit deviations from classical set-theoretical modeling, and traditional probabilistic approaches have difficulty coping with this. 
In this sense, the success of the quantum-theoretic modeling can be interpreted as a confirmation of the effectiveness of quantum theory to model conceptual combinations. We should also mention that our quantum-theoretic approach has recently produced new predictions, allowing us to identify entanglement in concept combinations \cite{as2011,as2014}, and systematic deviations from the marginal law, deeply connected to the mechanisms of concept formation \cite{asv2015,PhilTransA2015}. These effects would not have been identified in a more traditional investigation of overextension and underextension. 

It follows from the above analysis that our quantum-theoretic modeling rests  on a `theory based approach', as it straightforwardly derives from quantum theory as `a theory to represent natural concepts'.  Hence, it should be distinguished from an `ad hoc modeling based approach', only devised to fit data. One should be suspicious of models in which free parameters are added after the fact on an ad hoc basis to fit the data more closely. In our opinion, the fact that our `theory derived model' reproduces different sets of experimental data is a convincing argument to support its advantage over traditional modeling approaches and to extend its use to more complex combinations of concepts.

\section{The SCoP formalism as a contextual interfering prototype theory}\label{contextualprototype}
This section summarizes the SCoP approach to concepts by providing new insights to the research in \cite{ag2005a,ag2005b,gra2008}. 

We mentioned in Section \ref{intro} that, according to prototype theory, concepts are associated with a set of characteristic, rather than defining, features (or properties), that are weighted in the definition of the prototype. A new item is categorized as an instance of the concept if it is sufficiently similar to this prototype \cite{r1973,r1978,r1983}. 
The original prototype theory was subsequently put into mathematical form as follows.  The prototype consists of a set of features $\{ a_1, a_2, \ldots, a_M \}$, with associated `weights' (or `application values') $\{ x_{p1}, x_{p2}, \ldots, x_{pM} \}$, where $M$ is the number of features that are considered. A new item $k$ is also associated with a set $\{ x_{k1}, x_{k2}, \ldots, x_{kM} \}$, where the number $x_{km}$ refers to the applicability of the $m$-th feature to the item $k$ (for a given stimulus). Then, the conceptual distance between the item $k$ and the prototype, defined as
 \begin{equation}
 d_k=\sqrt{\sum_{m=1}^{M} (x_{km}-x_{pm})^{2}}
\end{equation}
is a measure of the similarity between item and prototype. The smaller the distance $d_k$ for the item $k$, the more representative $k$ is of the given concept. 

Prototype theory was developed in response to findings that people rate conceptual membership as graded (or fuzzy), with the degree of membership of an instance corresponding to the conceptual distance from the prototype. A second fundamental element of prototype theory is that it can in principle be confronted with empirical data, e.g., membership or typicality measurements.

A fundamental challenge to prototype theory (but also to any other theory of concepts) has become known as the `Pet-Fish problem'. The problem can be summarized as follows. We denote by {\it Pet-Fish} the conjunction of the concepts {\it Pet} and {\it Fish}. It has been shown that people rate {\it Guppy} neither as a typical {\it Pet} nor as a typical {\it Fish}, they do rate it as a highly typical {\it Pet-Fish} \cite{os1981}. This phenomenon of the typicality of a conjunctive concept being greater than 
 -- or overextends --
 that of either of its constituent concepts has also been called the `Guppy effect'. 
Using classical logic, or even fuzzy logic, there is no specification of a prototype for {\it Pet-Fish} starting from the prototypes of {\it Pet} and {\it Fish} that is consistent with empirical data \cite{z1982,os1981,os1982}. 
Fuzzy set theory falls short because standard connectives for conceptual conjunction involve typicality values that are less than or equal to each of the typicality values of the conceptual components, i.e. the typicality of an item such as {\it Guppy} is not higher for {\it Pet-Fish} than for either {\it Pet} or {\it Fish}.

Similar effects occur for membership weights of items with respect to concepts and their combinations. Hampton's experiments indicated that people estimate membership in such a way that the membership weight of an item for the conjunction (disjunction) of two concepts, calculated as the large number limit of relative frequency of membership estimates, is higher (lower) than the membership weight of this item for at least one constituent concept \cite{h1988a} (\cite{h1988b}). This phenomenon is referred to as `overextension' (`underextension'). `Double overextension' (`double underextension') is also an experimentally established phenomenon, when the membership weight with respect to the conjunction (disjunction) of two concepts is higher (lower) than the membership weights with respect to both constituent concepts \cite{h1988a} (\cite{h1988b}). Furthermore, conceptual negation does not satisfy the rules of classical Boolean logic \cite{h1997}. More,
Bonini, Osherson, Viale and Williamson \cite{bovw1999}, and then Alxatib and Pelletier 
 \cite{ap2011}, identified the presence of `borderline contradictions', directly connected with overextension, namely, a sentence like ``John is tall and John is not tall'' is estimated as true by a significant number of participants, again violating basic rules of classical Boolean logic. 
More generally, for each of these experimental data, a single classical probability framework satisfying the axioms of Kolmogorov does not exist \cite{a2009b,ags2013,a2009a,s2014a,s2014b,asv2015,abgs2013}. To clarify the latter sentence no single probability space can be constructed for an item whose membership weight with respect to the conjunction of two concepts is overextended with respect to both constituent concepts.

These problems -- compositionality, the graded nature of typicality, and the probabilistic nature of membership weights -- present a serious challenge to any theory of concepts.

We have developed a novel theoretical model of concepts and their combinations \cite{ga2002,ag2005a,ag2005b}, conjunction \cite{ags2013,a2009a,s2014a,s2014b,asv2015}, disjunction \cite{ags2013,a2009a}, conjunction and negation \cite{s2014b,asv2015}. It uses the mathematical formalism of quantum theory in Hilbert space to represent data on conceptual combinations, which has been successfully tested several times. This quantum-conceptual 
approach enables us to model the above-mentioned deviations from classicality in terms of genuine quantum phenomena (contextuality, emergence, entanglement, interference, and superposition), thus capturing fundamental aspects of how concepts combine. More importantly, we have recently  identified stronger deviations from classicality than overextension and underextension, which unveil, in our opinion, deep non-classical aspects of concept formation \cite{PhilTransA2015}. 

The approach was inspired by similarity based theories, such as prototype theory, in several respects:

(i) a fundamentally probabilistic formalism is needed to represent concepts and their dynamics;

(ii) the typicality of different items with respect to a concept is context-dependent;

(iii) features (or properties) of a concept vary in their applicability.

A key insight underlying our approach is considering a concept as, not a `container of instantiations' but, rather, an `entity in a specific state', which changes under the influence of a context. In our quantum-conceptual approach, a context is mathematically modeled as quantum physics models of a measurement on a quantum particle. The (cognitive) context changes the state of a concept in the way a measurement in quantum theory changes the state of a quantum particle \cite{ag2005a,ag2005b}. For example, in our modeling of the concept {\it Pet}, we considered the context $e$ expressed by {\it Did you see the type of pet he has? This explains that he is a weird person}, and found that when participants in an experiment were asked to rate different items of {\it Pet}, the scores for {\it Snake} and {\it Spider} were very high in this context. In this approach, this is explained by introducing different states for the concept {\it Pet}. We call `the state of {\it Pet} when no specific context is present' its ground state $\hat p$. The context $e$ changes the ground state $\hat p$ into a new state $p_{weird\ person\ pet}$. Typicality here is an observable semantic quantity, which means that it takes different values in different states of the concept. As a consequence, a substantial part of the typicality variations that are encountered in the Guppy effect are due to, e.g., changes of state of the concept {\it Pet} under the influence of a context. More specifically, the  typicality variations for the conjunction {\it Pet-Fish} are in great part similar to the typicality variations for {\it Pet} under the context {\it Fish} (and also for {\it Fish} under the context {\it Pet}). Not only does context play a role in shaping the typicality variations for {\it Pet-Fish}, but also interference between {\it Pet} and {\it Fish} contributes, as we will analyze in detail in Section \ref{quantummodel}.

In general, whenever someone is asked to estimate the typicality of {\it Guppy} with respect to the concept {\it Pet} in the absence of any context, it is the typicality in the ground state $\hat{p}_{Pet}$ that is obtained, and whenever the typicality of {\it Guppy} is estimated with respect to the concept {\it Fish} in the absence of any context, it is the typicality in the ground state $\hat{p}_{Fish}$ that is obtained. But, whenever someone is asked to estimate the typicality of {\it Guppy} with respect to the conjunction {\it Pet-Fish}, it is the typicality in a new ground state $\hat{p}_{Pet-Fish}$ that is obtained. This new ground state $\hat{p}_{Pet-Fish}$ is different from  
$\hat{p}_{Pet}$ as well as from $\hat{p}_{Fish}$. It is close but not equal to the changed state of the ground state $\hat{p}_{Pet}$ under the context $e_{Fish}$, and close but not equal to the changed state of the ground state $\hat{p}_{Fish}$ under the context $e_{Pet}$, the difference being due to interference taking place between {\it Pet} and {\it Fish} when they combine into {\it Pet-Fish} (see Section \ref{quantummodel}). The `changes of state under the influence of a context' and corresponding typicalities behave like the changes of state and corresponding probabilities behave in quantum theory', giving rise to a violation of corresponding fuzzy set and/or classical probability rules.
This partly explains the high typicality of {\it Guppy} in the conjunction {\it Pet-Fish}, and its normal typicality in {\it Pet} and {\it Fish}, and the reason why we identify the Guppy effect as an effect at least partly due to context. There is also an interference effect, as we will see later.

We developed this approach in a formal way, and called the underlying mathematical structure a `State Context Property (SCoP) formalism' \cite{ag2005a}. Let $A$ denote a concept. In SCoP, $A$ is associated with a triple of sets, namely the set $\Sigma$ of states -- we denote states by $p, q, \ldots$, the set ${\mathcal M}$ of contexts, we denote contexts by $e, f, \ldots$, and the set ${\mathcal L}$ of properties -- we denote properties by $a, b, \ldots$. The `ground state' $\hat{p}$ of the concept $A$ is the state where $A$ is not under the influence of any particular context. Whenever $A$ is under the influence of a specific context $e$, a change of the state of $A$ occurs. In case $A$ was in its ground state $\hat{p}$, the ground state changes to a state $p$. The difference between states $\hat{p}$ and $p$ is manifested, for example, by the typicality values of different items of the concept, as we have seen in the case of the Guppy effect, and the applicability values of different properties being different in the two states $\hat{p}$ and $p$. Hence, to complete the mathematical construction of SCoP, also two functions $\mu$ and $\nu$ are needed. The function $\mu: \Sigma \times {\mathcal M} \times \Sigma \longrightarrow [0, 1]$ is defined such that $\mu(q,e,p)$ is the probability that state $p$ of concept $A$ under the influence of context $e$ changes to state $q$ of concept $A$. The function $\nu: \Sigma \times {\mathcal L} \longrightarrow [0, 1]$ is defined such that $\nu(p,a)$ is the weight, or normalization of applicability, of property $a$ in state $p$ of concept $S$. The function $\mu$ mainly accounts for typicality measurements, the function $\nu$ mainly accounts for applicability measurements. Through these mathematical structures the SCoP formalism captures both `contextual typicality' and `contextual applicability' \cite{ag2005a}.

A quantum representation in a complex Hilbert space of data on {\it Pet} and {\it Fish} and different states of {\it Pet} and {\it Fish} in different contexts was developed \cite{ag2005a}, as well as of the concept {\it Pet-Fish} \cite{ag2005b}. Let us deepen the connections between the quantum-theoretic approach to concepts and prototype theory (see also \cite{gra2008}). This approach can be interpreted in a rather straightforward way  as a 
generalization of prototype theory which mathematically integrates context and formalizes its effects, unlike standard prototype theory.  What we call the ground state of a concept can be seen as the prototype of this concept. The conceptual distance of an item from the prototype can be reconstructed from the functions $\mu$ and $\nu$ in the SCoP formalism. Thus, as long as individual concepts are considered and in the absence of any context, prototype theory can be embodied into the SCoP formalism, and the prototype of a concept $A$ can be represented as its ground state $\hat{p_{A}}$. However, any context will change this ground state into a new state. An important consequence of this is that when the concept is in this new state, the prototype changes. An intuitive way of understanding this  is to consider this new state a new `contextualized prototype'. More concretely, the concept {\it Pet}, when combined with {\it Fish} in the conjunction {\it Pet-Fish}, has a new contextualized prototype,  which could be called `{\it Pet} contextualized by Fish'. The new state can be thought of as a `contextualized prototype'. Hence, this is a prototype-like theory that is capable of mathematically describing the presence and influence of context. From the point of view of conceptual distance, this contextualized prototype will be close to, e.g., {\it Guppy}.

The interpretation of the SCoP formalism as a contextual prototype theory can be applied not just to conjunctions and disjunctions of two concepts, but also to abstract categories such as {\it Fruits}. It is very likely that the prototype of {\it Fruits} is close to, e.g., {\it Apple}, or {\it Orange}. But let us now consider the combination {\it Tropical Fruits}, that is, {\it Fruits} under the context {\it Tropical}. It is then reasonable to maintain that the new contextualized prototype of {\it Tropical Fruits} is closer to, e.g., {\it Pineapple}, or {\it Mango}, than to {\it Apple}, or {\it Orange}. The introduction of contextualized prototypes within the SCoP formalism enables us to incorporate abstract categories as well as deviations of typicality from fuzzy set behavior.

Another interesting aspect of this approach to prototype theory comes to light if we consider again the conceptual combination {\it Pet-Fish}. It is reasonable that the prototypes of {\it Pet} and {\it Fish} -- ground states $\hat{p}_{Pet}$ and $\hat{p}_{Fish}$ -- interfere in {\it Pet-Fish} whenever the typicality of an item, e.g., {\it Guppy}, is measured with respect to {\it Pet-Fish}. This sentence cannot, however, be made more explicit in the absence
of a concrete quantum-theoretic representation of typicality measurements of items 
 with respect to concepts and their combinations. Indeed, interference and superposition  effects can be precisely formalized in such quantum representation. This will be the content of Sections \ref{quantummodel} and \ref{geometric}.

\section{A Hilbert space modeling of membership measurements}\label{quantummodel}
One can gain insight into how people combine concepts by gathering data on `membership weights' and `typicalities'. 
To obtain data on `typicalities', participants are given a concept, and a list of instances or items, and asked to estimate their typicality on a Likert scale. In other experiments 
participants are asked to choose which 
instance they consider most typical of the concept. Averages of these estimates or relative frequencies of the picked items
 give rise to values representing the typicalities of the respective items. A membership weight is obtained by asking participants to estimate the membership of specific items with respect to a concept. This estimation can be quantified using the 7-point Likert scale and then converted into a relative frequency, and then into a probability called the `membership weight'.

Hampton used membership weights instead of typicalities \cite{h1988b}, because all you can do with typicalities is fuzzy set type calculations: the minimum rule of fuzzy sets for conjunction or the maximum rule for disjunction. This approach has many serious shortcomings; indeed the Pet-Fish problem could not be addressed by it. More serious failures are revealed by membership weight data.  Hampton measured `membership weights' and `degrees of non-membership or membership', making these two measurements in one experiment. 
More specifically, Hampton's experiment generates magnitude data, 
measuring the
`degree of membership or non-membership' using a 7-point Likert scale providing $-1$, $-2$, $-3$ for degrees of 
non-membership, $1$, $2$, $3$ for degrees of 
membership and, $0$ for borderline cases. 
From the same experiment 
membership weight data are obtained, with 8 possible triplets $[\pm,\pm,\pm]$ per item. Each triplet indicating with a $+$ whether the participant considered item $k$ to be a member of the first category ($A$), the second category ($B$) and  the third disjunction category ($A \, {\rm or }\, B$), and with a $-$ respectively otherwise. 
In the present Hilbert space model we use the `degree of membership or non-membership' values obtained by Hampton, add +3 to them to make them all non-negative, sum them, and divide each one by this sum. 
Since there are 24 items in total, in this way we get a set of 24 values in the interval [0,1], that sum up to 1. We will use these values as a substitute for membership collapse probabilities.

Let us first explain how we arrive at the membership collapse probabilities as a consequence of a measurement, and why we can use the above-mentioned calculated values of `degree of membership or non-membership' as substitutes. Suppose that instead of using the data obtained by Hampton, we performed the following experiment. For each pair of concepts and their combination we ask the participant to select one and only one item that they consider the best choice for membership. Then we calculate for each of the 24 items the relative frequency of its appearance. These relative frequencies are 24 values in the interval [0,1] summing up to 1, and their limits for increasing numbers of participants represent the probabilities for each item to be chosen as the best member. These probabilities are what in a quantum model are called the `membership collapse probabilities'. Of course, the above described experiment to determine the membership collapse probabilities has not been performed. However, the values
calculated from Hampton's measurement of `degree of membership or non-membership', after renormalization as explained above, are expected to correlate with what these membership collapse probabilities would be if they were measured. This is why we use them as substitutes for the membership collapse probabilities in our quantum model. As we will see when we construct the quantum model, the exact values of the substitutes for the membership collapse probabilities are not critical. Thus, if we can model the substitutes for the membership collapse probabilities calculated from Hampton's data, we can also model the actual membership collapse probabilities (the data we would have if the experiment had been done). 

So, we repeat, in Table 1, Hampton's experimental data \cite{h1988b} have been converted into relative frequencies. The 
`degrees of non-membership and degrees of membership'  
give rise to $\mu_k(X)$ 
and now stand for the probability of concepts {\it Fruits} ($X=A$), {\it Vegetables} ($X=B$) and {\it Fruits or Vegetables} ($X$ = `$A$ or $B$') to collapse to the item $k$, and thus add up to 1, that is,
\begin{eqnarray}
\sum_{k=1}^{24}  \mu_{k} (A) = \sum_{k=1}^{24}  \mu_{k} (B) = \sum_{k=1}^{24}  \mu_{k} (A \ {\rm or}\ B) = 1  \label{normalizationcondition}
\end{eqnarray}
for the 24 items.
The quantum model for concepts and their disjunction in complex Hilbert space is developed by building appropriate state vectors and projection operators for a given ontology of 24 items of two more abstract `container' concepts.

\begin{table}[ht]
  \begin{center}
  \begin{tabular}{@{} clccccrrr @{}}
    \hline
               & &                      &                   A = FRUITS                         &      B = VEGETABLES      & &    &         \\
    \hline
   $k$  &  item & $ \mu_{k}(A)$  & $ \mu_{k}(B)$& $ \mu_{k}(A\ {\rm or}\ B)$& $\lambda_k$ & $\lambda$-rank &  $\epsilon_k$& $\phi_k$ \\ 
    \hline
    1 & Almond & 0.0359 & 0.0133 &0.0269 & 0.0217&16&+1&  84.0$^{\circ}$\\ 
    2 & Acorn & 0.0425 & 0.0108 &0.0249&0.0214&17&-1&-94.5$^{\circ}$ \\ 
    3 & Peanut &0.0372 & 0.0220 &0.0269&0.0285&10&-1&-95.4$^{\circ}$  \\ 
    4 & Olive & 0.0586& 0.0269 &0.0415 &0.0397&9&+1&91.9$^{\circ}$  \\ 
    5 & Coconut & 0.0755 & 0.0125 &0.0604 &0.0260&12&+1&57.7$^{\circ}$ \\ 
    6 & Raisin & 0.1026 &0.0170 &0.0555 &0.0415&7&+1&95.9$^{\circ}$  \\ 
    7 & Elderberry & 0.1138& 0.0170& 0.0480 &0.0404&8&-1&-113.3$^{\circ}$ \\ 
    8 & Apple & 0.1184 & 0.0155 &0.0688 &0.0428&5&+1&87.6$^{\circ}$   \\ 
    9 & Mustard & 0.0149 & 0.0250 &0.0146 &0.0186&19&-1 & -105.9$^{\circ}$ \\ 
   10 & Wheat & 0.0136 & 0.0255 &0.0165&0.0184&20&+1&99.3$^{\circ}$  \\ 
   11 & Root Ginger & 0.0157 & 0.0323 &0.0385 &0.0172&22&+1 &49.9$^{\circ}$ \\ 
    12 & Chili Pepper & 0.0167 & 0.0446 &0.0323&0.0272&11&-1 &-86.4$^{\circ}$ \\ 
   13 & Garlic & 0.0100 & 0.0301& 0.0293 &0.0146&23&-1 &-57.6$^{\circ}$\\ 
    14& Mushroom & 0.0140 & 0.0545 &0.0604 &0.0087&24&+1 &18.5$^{\circ}$  \\ 
   15 & Watercress & 0.0112 & 0.0658& 0.0482 &0.0253&13&-1 &-69.1$^{\circ}$ \\ 
   16 & Lentils & 0.0095 &0.0713& 0.0338&0.0252&14&+1& 104.7$^{\circ}$ \\ 
    17 & Green Pepper & 0.0324 & 0.0788 &0.0506&0.0503&4&-1 &-95.7$^{\circ}$\\ 
   18 & Yam & 0.0533 & 0.0724& 0.0541&0.0615&3&+1 & 98.1$^{\circ}$\\ 
   19 & Tomato & 0.0881 & 0.0679& 0.0688&0.0768&1&+1& 98.5$^{\circ}$\\ 
    20 & Pumpkin & 0.0797& 0.0713 &0.0579&0.0733&2&-1 & -103.5$^{\circ}$ \\ 
    21 & Broccoli & 0.0143 & 0.1284& 0.0642&0.0423&6&-1 & -99.5$^{\circ}$\\ 
    22 & Rice & 0.0140 & 0.0412 &0.0248 &0.0238&15&-1 & -96.7$^{\circ}$\\ 
   23 & Parsley & 0.0155 & 0.0266 &0.0308&0.0178&21&-1 &-61.1$^{\circ}$ \\ 
   24 & Black Pepper & 0.0127 & 0.0294 &0.0222&0.01929&18&+1 & 86.7$^{\circ}$ \\ 
    \hline
  \end{tabular}
  \end{center}
\vspace{1mm}
\noindent
 {\bf Table 1.} 
 Membership collapse probability values  $\mu_{k}(X)$ of 24 items for the categories  \emph{Fruits}, \emph{Vegetables} and, \emph{Fruits or Vegetables} \cite{h1988b}.  Notice also the 
 membership collapse probabilities for \emph{Mustard} and \emph{Pumpkin} still show the mark of double underextension of the disjunction.
Membership collapse probability data with $\delta \mu \approx 10^{-4}$ entail phase data $\delta \phi \approx 2 \cdot 10^{-1}$ and  lambda data $\delta \lambda \approx 4 \cdot 10^{-4}$.
\end{table}

In our model, the Hilbert space is a complex $n$-dimensional $\mathbb{C}^n$, in which state vectors are n-dimensional complex numbered vectors. We use the `bra-ket' notation  -- respectively $\langle \cdot \vert$  and $\vert \cdot   \rangle$ -- for vector states (see the Appendix for further explanation). The complex conjugate transpose of the $\vert \cdot  \rangle$  ket-vector (nx1 dim.) is the $\langle \cdot \vert$ bra-vector (1xn dim.). Projectors and operators are then combined as matrices $ \vert \cdot  \rangle \langle \cdot  \vert$, while scalars are obtained by inner product $ \langle \cdot  \vert \cdot   \rangle$.
We represent the measurement, consisting in the question ``Is item $k$ a good example of concept $X$?'', by means of an orthogonal projection operator $M_k$.  Each self-adjoint operator in the Hilbert space $\mathcal{H}$  has a spectral decomposition on $\{M_k \vert k = 1, \dots, 24\}$, where each $M_k$ is the projector corresponding to item $k$ from the list of 24 items in Table 1.
A priori we set no restrictions to the dimension of the complex Hilbert space, and thus neither to the projection space of the operators $M_k$.
Each separate concept \emph{Fruits}  and \emph{Vegetables} is now represented by its proper state vector $\vert A \rangle$ and $\vert B \rangle$ respectively, while their disjunction \emph{Fruits or Vegetables} is realized by their equiponderous superposition  $\frac{1}{\sqrt{2}} (\vert A \rangle +\vert B \rangle)$.
It is precisely this feature of the model -- its ability
to represent combined concepts as superposed states -- that
provides the interferential  composition of what could not be classically composed using sets.

Following the standard rule of average outcome values of quantum theory, the probabilities, $\mu_{k}(A)$, $\mu_{k}(B)$ and $\mu_{k}(A\ {\rm or} \ B)$ are given by:
\begin{eqnarray}
\mu_{k}(A)	&=&  \langle A|M_k|A\rangle \label{mua} \\ 
\mu_{k}(B)	&=&  \langle B |M_k| B\rangle  \label{mub} \\ 
\mu_{k}(A\ {\rm or} \ B)&=& \frac{\langle A|+  \langle B|}{\sqrt{2}}M_k\frac{|A\rangle +  |B\rangle}{\sqrt{2}} \label{states}
\end{eqnarray}
After a straightforward calculation, the
membership probability expression $\mu_{k}(A \ {\rm or} \ B)$ becomes:
 \begin{eqnarray}
 \mu_{k}(A\ \rm{or} \ B)	&=& \frac{1}{2} \left(\langle A| M_k |A\rangle+ \langle A| M_k |B\rangle + \langle  B| M_k |A\rangle +  \langle B| M_k |B\rangle\right) \nonumber \\
 		&=&  \frac{1}{2} \left(\mu_{k}(A) + \mu_{k}(B) \right) + \Re   \langle A| M_k |B\rangle  \label{muAorB}
 \end{eqnarray}
where $\Re$ takes the real part of  $\langle A| M_k |B\rangle$. This expression
shows the contribution  of the interference term $\Re\langle A|M_k|B\rangle$  in $ \mu_{k}(A\ \rm{or} \ B)$ with respect to the `classical average' term $ \frac{1}{2} \left(\mu_{k}(A) + \mu_{k}(B) \right)$. It consists of the real part of the complex probability amplitude of the k-th item in \emph{Vegetables} (concept $\vert B\rangle$) to be the one in \emph{Fruits} (concept $\vert A \rangle$).

The quantum concept model  imposes the orthogonality of the state vectors corresponding to different concepts. Therefore, we have for the states of \emph{Fruits} and \emph{Vegetables},
\begin{eqnarray} 
\langle A\vert B \rangle =0. \label{ABorthogonal}
\end{eqnarray}
Each different item of the projector $M_k$ also provides an orthogonal projection space. 
Since the conceptual disjunction \emph{Fruits or Vegetables} spans a subspace of 2 dimensions in the complex Hilbert space (along the rays of $\vert A \rangle$ and $\vert B \rangle$), we set forth the possibility for a complex 2-dimensional subspace for each item.
This brings the dimension of the complex Hilbert space to 48. However, we will choose the unit vectors of these subspaces in such a way as to eliminate redundant dimensions whenever possible. 
Each category vector is built on orthogonal unit vectors, defined by the projection operators $M_k$.  i.e. we define $|e_k\rangle$ the unit vector on $M_k|A\rangle$, and define  $|f_k\rangle$  the unit vector on $M_k|B\rangle$. Thus each item is now represented by a vector spanned by $|e_k\rangle$ and $|f_k\rangle$.   Due the orthogonality of the projectors $M_k$, we have
\begin{eqnarray} 
\langle e_k|f_l\rangle= \delta_{kl} c_k e^{i\gamma_k}  \label{innerproductofcomponents}
\end{eqnarray}
where the Kronecker
$\delta_{kl} = 1$ for same indices and zero otherwise, i.e. different item states are orthogonal as well. And $c_k$ expresses the angle between the two unit vectors $|e_k\rangle$ and $|f_k\rangle$ of each 2-dimensional subspace of item $k$. Notice that should some $c_k$ be 1, then the required dimension of the complex Hilbert space diminishes by 1, since the vectors $|e_k\rangle$ and $|f_k\rangle$ then coincide -- a property that we will use to minimize the size of the required Hilbert space. Should $c_k$ be different from 1, then $|e_k\rangle$ and $|f_k\rangle$ span a subspace of 2 dimensions.
The state vectors $ |A\rangle$ and $ |B\rangle$  of the concepts can then be expressed as a superposition of the vectors $|e_k\rangle$ and $|f_k\rangle$ for the items:  
\begin{eqnarray} 
|A\rangle=\sum_{k=1}^{24}a_ke^{i\alpha_k}|e_k\rangle, \ \ \ \ |B\rangle=\sum_{k=1}^{24} b_ke^{i\beta_k}|f_k\rangle 
\end{eqnarray}
where $a_k$, $b_k$, $\alpha_k$, $\beta_k \in \mathbb{R}$.

We can express their inner product as follows:
\begin{eqnarray} 
&\langle A|B\rangle=(\sum_{k=1}^{24}a_ke^{-i\alpha_k}\langle e_k|)(\sum_{l=1}^{24}b_le^{i\beta_l}|f_l\rangle) =\sum_{k=1}^{24}a_kb_kc_ke^{i(\beta_k-\alpha_k+\gamma_k)}  =\sum_{k=1}^{24}a_kb_kc_ke^{i\phi_k} \nonumber 
\end{eqnarray}
where we have defined phase $\phi_k$ as $\phi_k :=\beta_k-\alpha_k+\gamma_k$ in the last step.
The 
membership probabilities given in Eqs.
 (\ref{mua} and \ref{mub}) and the interference terms in Eq. (\ref{muAorB}) can be expanded on the projection spaces of the items:
\begin{eqnarray}
\mu_{k}(A) &=& (\sum_{l=1}^{24}a_le^{-i\alpha_l}\langle e_l|)(a_ke^{i\alpha_k}|e_k\rangle)=a_k^2 \\
\mu_{k}(B) &=& (\sum_{l=1}^{24}b_le^{-i\beta_l}\langle f_l|)(b_ke^{i\beta_k}|f_k\rangle)=b_k^2 \\
\langle A|M_k|B\rangle &=& (\sum_{l=1}^{24}a_le^{-i\alpha_l}\langle e_l|)M_k|(\sum_{m=1}^{24}b_me^{i\beta_m}|f_m\rangle) = a_kb_ke^{i(\beta_k-\alpha_k)}\langle e_k|f_k\rangle=a_kb_kc_ke^{i\phi_k} \label{interferenceterm}
\end{eqnarray}
Notice that the phase of the $k$-th component  of the conceptual disjunction  is not at play in the interference term $\langle A\vert M_k \vert B\rangle$ (Eq. \ref{muAorB}).
Taking the real part of the interference term in Eq. (\ref{interferenceterm}), we can rewrite the
membership probability of the disjunction in Eq. (\ref{muAorB}) as follows:
 \begin{eqnarray}
 \mu_{k}(A\ \rm{or} \ B)&=& \frac{1}{2} \left(\mu_{k}(A) + \mu_{k}(B) \right) + c_k \sqrt{ \mu_{k}(A) \mu_{k}(B)} \cos \phi_k
  \end{eqnarray}
 Rearranging this equation we now choose $\phi_k$ must satisfy
\begin{eqnarray}
\cos\phi_k = \frac{\mu_{k}(A\ {\rm or}\ B) - {1 \over 2}(\mu_{k}(A)+\mu_{k}(B))}{c_k\sqrt{\mu_{k}(A)\mu_{k}(B)}} \label{composedtypicalitycos}
\end{eqnarray}
Since all the membership probabilities on the right side are fixed, the only remaining free parameters are the coefficients $c_k$. These parameters must now be tuned in order to satisfy the  orthogonality of $\vert A \rangle$ and  $\vert B \rangle$. Using the expansion on the unit vector sets $\{\vert e_k \rangle\}$, $\{\vert f_k \rangle\}$ we obtain for their orthogonality  Eq. (\ref{ABorthogonal}):
\begin{eqnarray}
\sum_{k=1}^{24}  c_k \sqrt{\mu(A)_k\mu(B)_k}\cos\phi_k &=&0, \label{cosinesum}\\
\sum_{k=1}^{24} c_k \sqrt{\mu(A)_k\mu(B)_k}\sin\phi_k &=&0. \label{sinesum}
\end{eqnarray}
The `cosine sum' (Eq. \ref{cosinesum}) is automatically satisfied due to the definition of $\cos \phi_k$ and the normalisation of
membership probabilities (Eq. \ref{normalizationcondition}). 
This can be seen by substituting the expression of $\cos \phi_k$ in Eq. (\ref{composedtypicalitycos}) and then applying the normalization condition  of the
membership probabilities
(Eq. \ref{normalizationcondition}). The `sine sum' equation still needs to be satisfied. With the defining relation Eq. (\ref{composedtypicalitycos}) of $\phi_k$, and $\sin \phi_k = \epsilon_k \sqrt{1-\cos^2 \phi_k}$, where $\epsilon_k=\pm$ provides the sign, this becomes\footnote{The cosine value only defines the phase up to its absolute value $\vert \phi_k\vert$. Thus the sign of the sine value is undefined. If  $\epsilon_k = -1$, then $\phi_k = - \vert \phi_k\vert$.}
\begin{eqnarray}
\sum_{k=1}^{24} \epsilon_k \sqrt{c_k^2 \mu_{k}(A)\mu_{k}(B) - (\mu_{k}(A\ {\rm or}\ B) - {1 \over 2}(\mu_{k}(A)+\mu_{k}(B)) )^2} =0. \label{sinesumbis}
\end{eqnarray}
In order to satisfy this equation a simple algorithm was devised \cite{a2009a}. For convenience of notation we denote the square root expression, with $c_k =1$, by a separate symbol:
\begin{eqnarray}
\lambda_k :=  \sqrt{\mu_{k}(A)\mu_{k}(B) - (\mu_{k}(A\ {\rm or}\ B) - {1 \over 2}(\mu_{k}(A)+\mu_{k}(B)) )^2} . \label{definelambda}
\end{eqnarray}
 First, we order the values $\lambda_k$ from large to small and then assign a sign $\epsilon_k$ to each of them in such a way that each next partial sum (increasing index) remains smallest. The $\lambda$-ranking with corresponding values have been tabulated in Table 1.
 We assign index $m$ to the item with the largest $\lambda$-value. In the present case, the item \emph{Tomato} has the largest  value, 0.07679.
  
 We  now adopt an optimized complex Hilbert space for our model in which $c_k =1$ ($k \neq m$), which reduces the space to 25 complex dimensions. We again note
 that 
 all items except {\it Tomato} receive a 1-dimensional complex subspace, while {\it Tomato} is
  represented by a 2-dimensional subspace.
The `sine sum' equation in Eq. (\ref{sinesumbis}) 
can be written as
\begin{eqnarray}
\sum_{k=1, k\neq m}^{24} \epsilon_k \lambda_k  +  \epsilon_m \sqrt{c_m^2 \mu_{m}(A)\mu_{m}(B) - (\mu_{m}(A\ {\rm or}\ B) - {1 \over 2}(\mu_{m}(A)+\mu_{m}(B)) )^2} =0.
\end{eqnarray}
Next, we define the partial sum of the $\lambda_k$ according a scheme of signs $\epsilon_k$ such that from large to small the next $\epsilon_k \lambda_k$ is added to make the sum smaller but not negative. 
\begin{eqnarray}
S_j &=& \sum_{size\ ordered\  \lambda_i}^j \epsilon_i \lambda_i   \label{partiallambdasum} \\  
    S_{j+1} &=& S_j - \lambda_{j+1}  \  {\rm and} \   \epsilon_{j+1} = -1,   \  {\rm if} \   S_{j}  - \lambda_{j+1} \geq 0 \\
 &=& S_j + \lambda_{j+1}  \  {\rm and} \  \epsilon_{j+1} = +1, \  {\rm if} \   S_{j}  - \lambda_{j+1} < 0 
\end{eqnarray}
The first summand is thus $\lambda_m$, with $\epsilon_m =+1$. Finally this procedure leads to 
\[
S_{24} = \sum_{k=1}^{24} \epsilon_k \lambda_k \geq 0 
\]
In the
 \emph{Fruits} and \emph{Vegetables} example with
membership probability data in Table 1, this  procedure gives:
\begin{eqnarray}
S_{24 } &=& 0.0154  \label{S24}
\end{eqnarray}
In general the `sine sum' equation then becomes
\begin{eqnarray}
S_{24} - \lambda_m +  \sqrt{c_m^2 \mu_{m}(A)\mu_{m}(B) - (\mu_{m}(A\ {\rm or}\ B) - {1 \over 2}(\mu_{m}(A)+\mu_{m}(B)) )^2} = 0. \label{S24equation}
\end{eqnarray}
From which we can fix $c_m$, the remaining $c_k$ not equal to 1:
\begin{eqnarray}
c_m   &=& \sqrt{ \frac{ (S_{24} -   \lambda_m)^2 +  (\mu_{m}(A\ {\rm or}\ B) - {1 \over 2}(\mu_{m}(A)+\mu_{m}(B)) )^2 }{ \mu_{m}(A)\mu_{m}(B)} }
\end{eqnarray}
In the present example we obtain the value $c_m = 0.8032$. We thus have fixed the inner product  -- or `angle' -- of the vectors $|e_m \rangle$ and $|f_m\rangle$, and can now write an explicit representation in the canonical 25 dimensional complex Hilbert space $\mathbb{C}^{25}$. We can take $M_k(\mathcal{H})$ to be rays of dimension 1 for $k \neq m$, and $M_m(\mathcal{H})$ to be a 2-dimensional plane spanned by the vectors $|e_m \rangle$ and $|f_m\rangle$.

We let the space $\mathbb{C}^{25}$ be spanned on the canonical base $\{  \mathbf{1}_i \}$, $i \in [1\dots 25]$. All items $k\neq m$ are represented by the respective $ \mathbf{1}_i$. While for $k=m$ we express the projections  of  $|A\rangle$ and $|B\rangle$ by $M_m(\mathcal{H})$ accordingly on 
$\mathbf{1}_m$ and $\mathbf{1}_{25}$
\begin{eqnarray}
a_m e^{ i \alpha_m} \vert e_m\rangle &=& \tilde a_m e^{ i \alpha_{m_1}} \mathbf{1}_m + \tilde a_{25} e^{ i \alpha_{m_2}}  \mathbf{1}_{25} \\
b_m e^{ i \beta_m}  \vert f_m\rangle &=& \tilde b_m e^{ i \beta_{m_1}}  \mathbf{1}_m + \tilde b_{25}  e^{ i \beta_{m_2}}  \mathbf{1}_{25} 
\end{eqnarray}
with $\tilde a_m$, $\tilde b_m$, $\tilde a_{25}$, $\tilde b_{25} \in \mathbb{R}$ to be specified.  The parameters in these expressions should satisfy the inner product Eq. (\ref{innerproductofcomponents}) for $k, l=m$
\begin{eqnarray}
 a_m  b_m \langle     e_m    \vert f_m\rangle &=& \tilde a_m \tilde b_m  e^{- i (\alpha_{m_1}-\beta_{m_1})}  + \tilde a_{25}  \tilde b_{25} e^{- i (\alpha_{m_2}-\beta_{m_2})} ,  \\
 &=&  a_m  b_m  c_m e^{i (\gamma_m -\alpha_m+\beta_m)} 
\end{eqnarray}
and the probality weights for $k=m$
\begin{eqnarray}
a_m^2   &=& \tilde a_m^2 + \tilde a_{25}^2, \\
b_m^2   &=& {\tilde b}_m^2 + {\tilde b}_{25}^2 .
\end{eqnarray}
Finally, the representation of all vectors of the items can now  be rendered explicit by simply choosing $\alpha_k = \gamma_k=0$, and thus $\beta_k = \phi_k$, $\forall k$. A further simplification for 
{\it Tomato} is done by setting $\tilde a_{25}=0$, which also allows free choice of $\beta_{m_2}=0$. Then 
$  \tilde a_m = a_m  $  and $  \tilde b_m = b_m c_m $, and ${\tilde b}_{25} = b_m \sqrt{ (1-c_m^2)} $.
We have rendered explicit these
membership probabilities and phases in Table 1.
Thus we can 
write 
the vectors $\vert A\rangle$ and $\vert B \rangle$ in $\mathbb{C}^{25}$ Hilbert space corresponding to the categories \emph{Fruits} and \emph{Vegetables}  respectively.
\begin{eqnarray} \label{interferenceangles01}
&|A\rangle=(
0.1895,
0.2062,
0.1929,
0.2421,
0.2748,
0.3203,
0.3373,
0.3441,
0.1221, \nonumber \\
&0.1166,
0.1253,
0.1292,
0.1000,
0.1183,
0.1058,
0.0975,
0.1800,
0.2309, \nonumber\\
&0.2968,
0.2823,
0.1196,
0.1183,
0.1245,
0.1127,
0.0000)  \\ \label{interferenceangles02}
&|B\rangle=(
0.1153e^{i 84.0^{\circ}},
0.1039e^{-i 94.5^{\circ}},
0.1483e^{-i 95.4^{\circ}},
0.1640e^{i 91.9^{\circ}},\nonumber \\
&
0.1118e^{i 57.7^{\circ}},
0.1304e^{i 95.9^{\circ}},
0.1304e^{-i 113.3^{\circ}},
0.1245e^{i 87.6^{\circ}},\nonumber \\
&
0.1581e^{-i 105.9^{\circ}},
0.1597e^{i 99.3^{\circ}},
0.1797e^{i 49.9^{\circ}},
0.2112e^{-i 86.4^{\circ}},\nonumber \\
&
0.1735e^{-i 57.6^{\circ}},
0.2335e^{i 18.5^{\circ}},
0.2565e^{-i 69.1^{\circ}},
0.2670e^{i 104.7^{\circ}},\nonumber \\
&
0.2807e^{-i 95.7^{\circ}},
0.2691e^{i 98.0^{\circ}},
0.2606e^{i 96.8^{\circ}},
0.2670e^{-i 103.5^{\circ}},\nonumber \\
&
0.3583e^{-i 99.5^{\circ}},
0.2030e^{-i 96.7^{\circ}},
0.1631e^{-i 61.1^{\circ}},
0.1715e^{i 86.7^{\circ}},\nonumber \\
&
0.1552
).
\end{eqnarray}
This completes the quantum model for the membership probability of items with respect to {\it Fruits}, {\it Vegetables} and {\it Fruits or Vegetables}. It captures the enigmatic aspects of conceptual overextension and underextension identified in \cite{h1988b}, explaining them in terms of
genuine quantum phenomena.

Recalling the terminology adopted in Section \ref{contextualprototype}, the unit vectors $|A\rangle$ and $|B\rangle$ in Eqs. (\ref{interferenceangles01}) and (\ref{interferenceangles02}) represent the ground states of the concepts {\it Fruits} and {\it Vegetables}, respectively. Equivalently, these unit vectors represent the prototypes of the concepts {\it Fruits} and {\it Vegetables} in prototype theory. The unit vector $\frac{1}{\sqrt{2}}(|A\rangle+|B\rangle)$ instead represents the `contextualized prototype' obtained by combining the prototypes of {\it Fruits} and {\it Vegetables} in the disjunction {\it Fruits or Vegetables}.  If one now looks at Eq. (\ref{muAorB}), one sees that the prototypes {\it Fruits} and {\it Vegetables} interfere in the disjunction {\it Fruits or Vegetables}, and the term $\Re\langle A|M_k|B\rangle$ in Eq. (\ref{muAorB}) specifies how much interference is present when the membership probability of $k$ is measured.

\section{An illustration of interfering prototypes}\label{geometric}
In this section we provide  an illustration of contextual interfering prototypes. It is not a complete mathematical representation as presented in Section \ref{quantummodel} but, rather, an illustration that can help the reader with a non-technical background to have an intuitive picture of what a contextual prototype is and how contextual prototypes interfere. Consider the concepts {\it Fruits}, {\it Vegetables} and their disjunction {\it Fruits or Vegetables}. The contextual prototype of {\it Fruits} can be represented by the x-axis of a plane surrounded by a cloud containing items, features, etc. -- all the contextual elements connected with the prototype of {\it Fruits}. Similarly, the contextual prototype of {\it Vegetables} can be represented by the y-axis of the same plane surrounded by a cloud containing items, features, etc. -- all the contextual elements connected with the prototype of {\it Vegetables}. How can we represent the contextual prototype of the disjunction {\it Fruits or Vegetables}?  Although as we have seen it cannot be represented in traditional fuzzy set theory, it can be represented in terms of waves, with peaks and troughs.  Indeed, waves can be summed up in such a way that peaks and troughs of the combined wave reproduce overextension and underextension of the data. In other words, waves 
provide an intuitive geometric illustration of the interference taking place when contextual prototypes are combined in concept combination as discussed in Section \ref{quantummodel}.
For  example, let us demonstrate the interference of the item 
\emph{Almond} when its  membership probability with respect to the disjunction {\it Fruits or Vegetables} is
calculated based on its
membership probabilities for  {\it Fruits} and for   {\it Vegetables}. The
membership probabilities for the categories  {\it Fruits}, {\it Vegetables} and {\it Fruits or Vegetables} have been calculated from the Hampton's data and are reported in Table 1.

The idea of an illustration would be to show that in addition to  `fuzziness'
(as modeled using a fuzzy set-theoretic approach) there is a `wave structure'. How can we graphically represent this `wave structure' of a concept? We start from the standard interference formula of quantum theory, which is the following. For an arbitrary item $k$ we have
\begin{eqnarray}
\mu_k(A\ {\rm or}\ B)={1 \over 2}(\mu_k(A)+\mu_k(B))+c_k\sqrt{\mu_k(A)\mu_k(B)}\cos\phi_k .
\end{eqnarray}
Now, we have
\begin{eqnarray}
\phi_k=\beta_k-\alpha_k+\gamma_k
\end{eqnarray}
where $\alpha_k$ is the phase angle connected with $\mu_k(A)$, $\beta_k$ the phase angle connected to $\mu_k(B)$, and $\gamma_k$ the phase angle connected to $\langle A|M_k|B\rangle$. This has not yet been emphasized but if one analyses the rest of the construction in Hilbert space, it is possible to see that one can always choose $\gamma_k=0$, which means that, with this choice, $\phi_k$ becomes the difference in phases $\beta_k$ and $\alpha_k$.

This is all we need to represent the `wave' nature of a concept in a manner analogous to that
of quantum theory. Indeed, it is the `phase difference' between the waves -- their phases being $\alpha_k$ and $\beta_k$ respectively -- that we attach to $\mu(A)_k$ and $\mu(B)_k$. They determine, together with the 
membership probabilities $\mu(A)_k$ and $\mu(B)_k$ 
the interference
that gives rise to the measured data for $\mu(A\ {\rm or}\ B)_k$.
 
The choice of the $c_k$ is such that only for the biggest value of $\lambda_k$, which in this case of {\it Tomato}, the $c_k$ is chosen different from 1. The only choice different from 1, for {\it Tomato}, still does not influence the fact that $\phi_k$ is the difference between $\beta_k$ and $\alpha_k$, when we decide to choose $\gamma_k=0$. Let us consider for example the first item {\it Almond} of the list of 24 in Table 1. We have 
\begin{eqnarray}
\mu(A)_1=0.0359 \\
\mu(B)_1=0.0133 \\
\mu(A\ {\rm or}\ B)_1=0.0269
\end{eqnarray}
These are the data measured by Hampton, and also what exists for the concepts {\it Fruits}, {\it Vegetables} and their combination {\it Fruits or Vegetables} with respect to membership probability of the item {\it Almond} in the realm where fuzzy set probability appears. These are the values that do not fit into a model in this realm, and for which a wave-like realm underneath is necessary. Calculating the angle $\phi_1$ we get
\begin{eqnarray}
\phi_1=84.0^\circ
\end{eqnarray}
(see Table 1).
This angle is the result of a wave being present underneath the fuzzy, 
probability realm for $\mu(A)_1$ and $\mu(B)_1$, such that both waves give rise to a difference in phase --  where the crests 
of one wave meet the troughs of the other -- which is equal to $\beta_1-\alpha_1$, and is the value of $\phi_1$.
This can be represented graphically by attaching a wave pattern to $\mu(A)_1$ and another one to $\mu(B)_1$, such that both have a phase difference of $84.0^\circ$ -- see also Figure 1.

\begin{figure}[ht] 
\centering
\includegraphics[width=7cm]{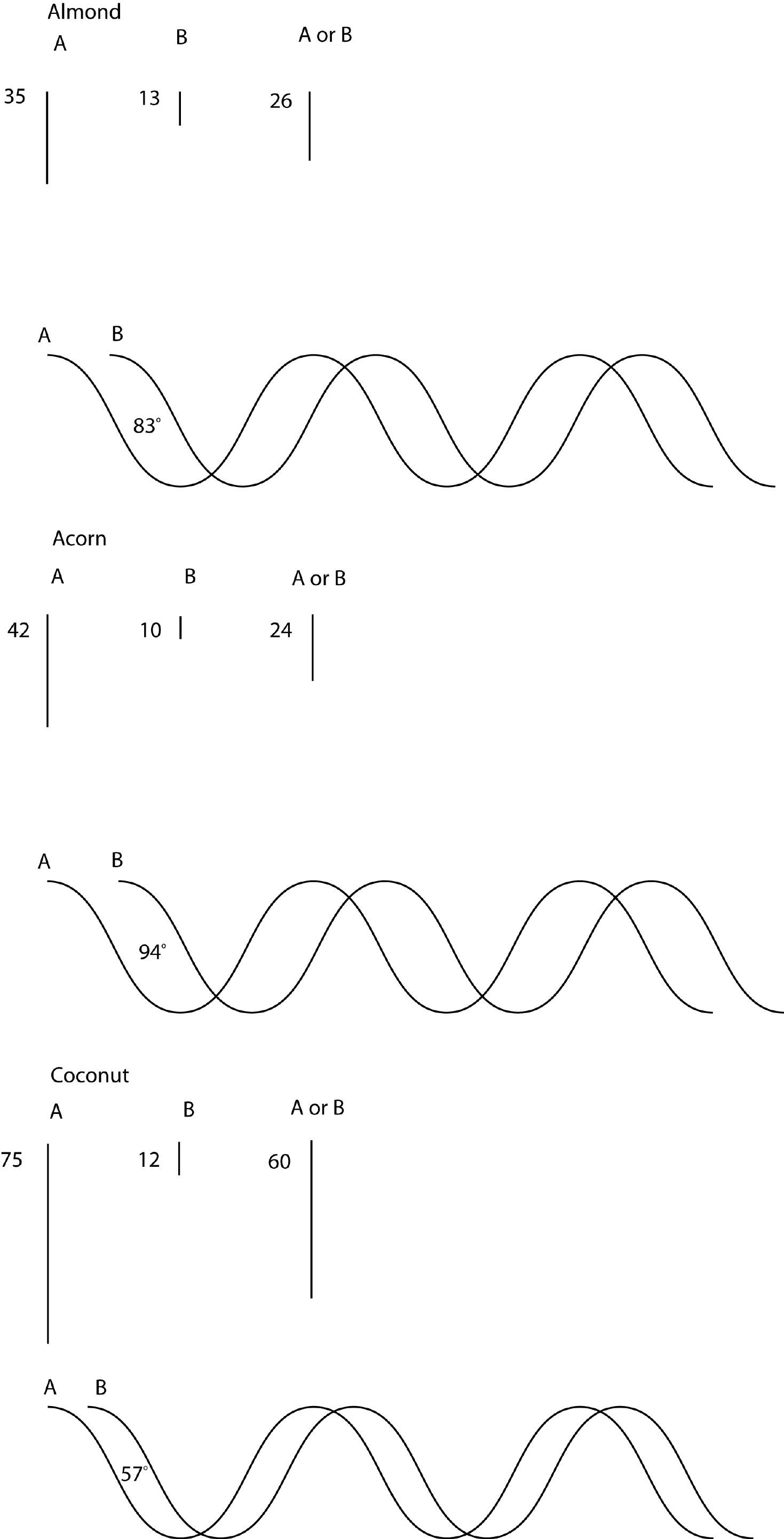}
\caption{Interference of items {\it Almond}, {\it Acorn} and {\it Coconut} in the concept \emph{Fruits or Vegetables}.  Elementary oscillatory waves $\sqrt{\mu_k (A)} \cos(x)$ and $\sqrt{\mu_k (B)} \cos(x + \phi_k)$ are associated to the components of each given item in \emph{Fruits} and \emph{Vegetables} respectively. The weight amplitude of the item in the disjunction \emph{Fruits or Vegetables} emerges at the origin of  $(\sqrt{\mu_k (A)} \cos(x)+ \sqrt{\mu_k (B)} \cos(x + \phi_k))/\sqrt{2}$.}
\end{figure}

Let us apply quantum theory to each of the items apart. 
Each item $k$ has a Schr\"odinger wave function vibrating in the neighborhood of {\it A}, another one vibrating in the neighborhood of {\it B} and a third vibrating in the neighborhood of `$A \ {\rm or} \ B$', and they are related by
superposition.
We have:
\begin{eqnarray}
\psi^A_k=\sqrt{\mu_k(A)}e^{i\alpha_k} \\
\psi^B_k=\sqrt{\mu_k(B)}e^{i\beta_k} \\
\psi^{A {\rm or} B}_k=\sqrt{\mu_k(A\ {\rm or}\ B)}e^{i\delta_k}
\end{eqnarray} 
In each case, this gives us the membership probabilities. Squaring (multiplying by its complex conjugate), we have
\begin{eqnarray}
&&\langle \psi^A_k|\psi^A_k\rangle=(\psi^A_k)^*(\psi^A_k)=(\sqrt{\mu_k(A)}e^{i\alpha_k})^*(\sqrt{\mu_k(A)}e^{i\alpha_k}) \nonumber \\
&&=(\sqrt{\mu_k(A)}e^{-i\alpha_k})(\sqrt{\mu_k(A)}e^{i\alpha_k})=\mu_k(A)e^{i(\alpha-\alpha)}=\mu_k(A) \\
&&\langle \psi^B_k|\psi^B_k\rangle=(\psi^B_k)^*(\psi^B_k)=(\sqrt{\mu_k(B)}e^{i\beta_k})^*(\sqrt{\mu_k(B)}e^{i\beta_k}) \nonumber \\
&&=(\sqrt{\mu_k(B)}e^{-i\beta_k})(\sqrt{\mu_k(B)}e^{i\beta_k})=\mu_k(B)e^{i(\beta-\beta)}=\mu_k(B) \\
&&\langle \psi^{A {\rm or} B}_k|\psi^{A {\rm or} B}_k\rangle=(\psi^{A {\rm or} B}_k)^*(\psi^{A {\rm or} B}_k)=(\sqrt{\mu_k(A\ {\rm or}\ B)}e^{i\delta_k})^*(\sqrt{\mu_k(A\ {\rm or}\ B)}e^{i\delta_k}) \nonumber \\
&&=(\sqrt{\mu_k(A\ {\rm or}\ B)}e^{-i\delta_k})(\sqrt{\mu_k(A\ {\rm or}\ B)}e^{i\delta_k})=\mu_k(A\ {\rm or}\ B)e^{i(\delta-\delta)}=\mu_k(A\ {\rm or}\ B) 
\end{eqnarray} 

If we write the quantum superposition equation for each item we get
\begin{eqnarray} \label{superposition}
&&{1 \over \sqrt{2}}(\psi^A_k+\psi^B_k)=\psi^{A {\rm or} B}_k \\
&\Leftrightarrow& {1 \over \sqrt{2}}(\sqrt{\mu(A)_k}e^{i\alpha_k}+\sqrt{\mu(B)_k}e^{i\beta_k})=\sqrt{\mu(A\ {\rm or}\ B)_k}e^{i\delta}_k
\end{eqnarray}
where ${1 \over \sqrt{2}}$ is a normalization factor. It is the squaring (i.e. multiplying each with its complex conjugate) that
 gives rise to the interference equation. Let us do this explicitly to see it. First we multiply the left hand side with its complex conjugate.
We do the multiplication explicitly writing each step of it, to see well how the interference formula appears. Hence, we have
\begin{eqnarray}
&&({1 \over \sqrt{2}}(\sqrt{\mu_k(A)}e^{i\alpha_k}+\sqrt{\mu_k(B)}e^{i\beta_k}))^*({1 \over \sqrt{2}}(\sqrt{\mu_k(A)}e^{i\alpha_k}+\sqrt{\mu_k(B)}e^{i\beta_k})) \nonumber \\
&=&({1 \over \sqrt{2}}(\sqrt{\mu_k(A)}e^{-i\alpha_k}+\sqrt{\mu_k(B)}e^{-i\beta_k}))({1 \over \sqrt{2}}(\sqrt{\mu_k(A)}e^{i\alpha_k}+\sqrt{\mu_k(B)}e^{i\beta_k})) \nonumber \\
&=&{1 \over 2}(\sqrt{\mu_k(A)}e^{-i\alpha_k}+\sqrt{\mu_k(B)}e^{-i\beta_k})(\sqrt{\mu_k(A)}e^{i\alpha_k}+\sqrt{\mu_k(B)}e^{i\beta_k})) \nonumber \\
&=&{1 \over 2}(\sqrt{\mu_k(A)}e^{-i\alpha_k} \cdot \sqrt{\mu_k(A)}e^{i\alpha_k}+\sqrt{\mu_k(A)}e^{-i\alpha_k} \cdot \sqrt{\mu_k(B)}e^{i\beta_k} \nonumber \\
&&+ \sqrt{\mu_k(B)}e^{-i\beta_k} \cdot \sqrt{\mu_k(A)}e^{i\alpha_k}+\sqrt{\mu_k(B)}e^{-i\beta_k} \cdot \sqrt{\mu_k(B)}e^{i\beta_k}) \nonumber \\
&=&{1 \over 2}(\mu_k(A)e^{i(\alpha_k-\alpha_k)}+\sqrt{\mu_k(A)\mu_k(B)}e^{i(\beta_k-\alpha_k)}+\sqrt{\mu_k(A)\mu_k(B)}e^{-i(\beta_k-\alpha_k)}+\mu_k(B)e^{i(\beta_k-\beta_k)}) \nonumber 
\end{eqnarray}
we use now that $e^{i(\alpha_k-\alpha_k)}=e^0=1$, $e^{i(\beta_k-\beta_k)}=e^0=1$, $e^{i(\beta_k-\alpha_k)}=\cos(\beta_k-\alpha_k)+i\sin(\beta_k-\alpha_k)$ and $e^{-i(\beta_k-\alpha_k)}=\cos(\beta_k-\alpha_k)-i\sin(\beta_k-\alpha_k)$, to get to the following
\begin{eqnarray}
&=&{1 \over 2}(\mu_k(A)+\sqrt{\mu_k(A)\mu_k(B)}(\cos(\beta_k-\alpha_k)+i\sin(\beta_k-\alpha_k)) \nonumber \\
&&+\sqrt{\mu_k(A)\mu_k(B)}(\cos(\beta_k-\alpha_k)-i\sin(\beta_k-\alpha_k))+\mu_k(B)) \nonumber 
\end{eqnarray}
see that the term in $i\sin(\beta_k-\alpha_k)$ cancels, to get
\begin{eqnarray}
&=&{1 \over 2}(\mu_k(A)+2\sqrt{\mu_k(A)\mu_k(B)}\cos(\beta_k-\alpha_k)+\mu_k(B)) \nonumber \\
&=&{1 \over 2}(\mu_k(A)+\mu_k(B))+\sqrt{\mu_k(A)\mu_k(B)}\cos(\beta_k-\alpha_k)
\end{eqnarray}
Let is multiply now the right hand sight of equation (\ref{superposition}) with its complex conjugate. This gives
\begin{eqnarray}
&=&(\sqrt{\mu_{k}(A\ {\rm or}\ B)}e^{i\delta}_k)^*(\sqrt{\mu_{k}(A\ {\rm or}\ B)}e^{i\delta}_k)=\mu_{k}(A\ {\rm or}\ B)
\end{eqnarray}
Hence, we get, as a consequence of squaring Eq. (\ref{superposition}), exactly our interference formula 
\begin{eqnarray}
{1 \over 2}(\mu_{k}(A)+\mu_{k}(B))+\sqrt{\mu_{k}(A)\mu_{k}(B)}\cos(\beta_k-\alpha_k)=\mu_{k}(A\ {\rm or}\ B)
\end{eqnarray}

Note that the difference in phase $\beta_k-\alpha_k$ between the waves connected with the item
 $k$ and {\it A} and the item 
 $k$ and {\it B} is what generates  the interference. The new wave connected to the item $k$ and {\it A or B}, of which the phase is $\delta$ is not influenced by it, is the amplitude of this new wave which is affected. This is the reason that interference is visible in the realm where the fuzzy nature appears, while it is provoked by the realm where the waves 
occur.

We put forward this `wave nature' aspect of concepts not just as an illustration, but to help the reader understand the manner in which such an underlying wave structure increases substantially the possible ways in which concepts can interact, as compared to 
the interaction possibilities in a modeling with fuzzy set structures.  Of course the notion of a `wave' only adds clarification if we can imagine it to exist in some space-like realm. This is the case for the type of waves we all know from our daily physical environment, such as water waves, sound waves or light waves. The quantum waves of physical quantum particles can also be made visible in general by looking at probabilistic detection patterns of these quantum particles on a physical screen, and noting the typical interference patterns when the waves interact and the particles are detected on the screen. One might believe that an analogous situation is not possible for concepts, because intuitively concepts, unlike quantum particles, do not exist `inside' space. If we look at things is an operational way, however, an analysis can be put forward for the quantum model of the combination of the two concepts, and the graded structure of collapse probability weights of the 24 items, which does illustrate the presence of an interference pattern, and as a consequence reveals the underlying wave structure of concepts and their interactions. Let us explain how we can proceed to accomplish such an analysis.

\begin{figure}[ht] 
\centering
\includegraphics[width=15cm]{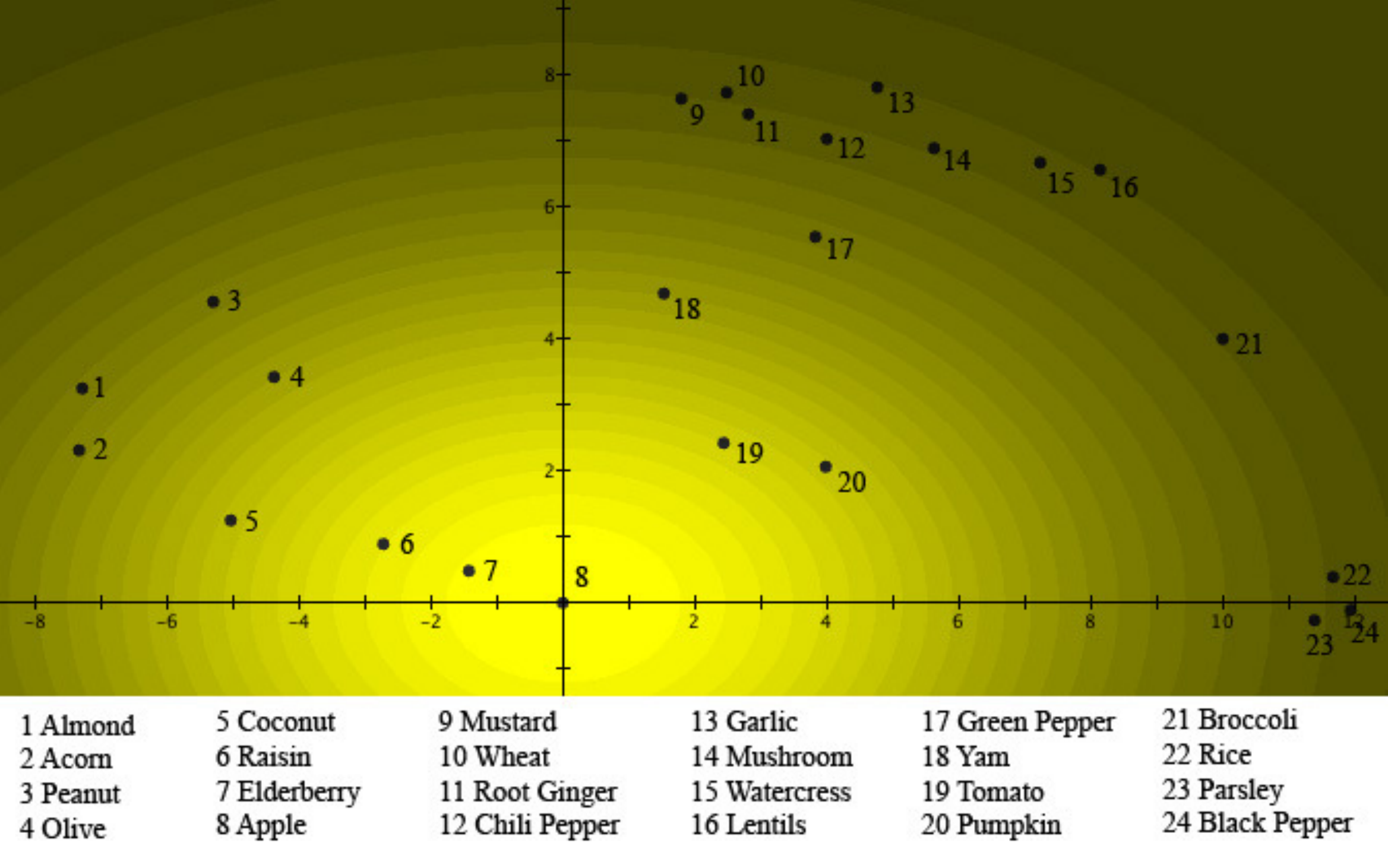}
\caption{The probabilities $\mu(A)_k$ of a person choosing the item $k$ as a `good example' of {\it Fruits} are fitted into a two-dimensional quantum wave function $\psi_A(x,y)$. The numbers are placed at the locations of the different items with respect to the Gaussian probability distribution $|\psi_A(x,y)|^2$. This can be seen as a light source shining through a hole centered on the origin, and regions where the different items are located. The brightness of the light source in a specific region corresponds to the probability that this item will be chosen as a `good example' of {\it Fruits}.} 
\end{figure}

We start by considering Figure 2. We see there the 24 different items of Table 1 represented by numbered spots in a plane where a graded pattern, starting with the lightest region around the spot number 8, which is {\it Apple}, systematically becomes darker. Different numbers of items are situated in spots in regions of different darkness, for example, number 16, {\it Lentils}, is situated in a spot in the 
darkest region. Let us explain how the figure is constructed. The `intensity of light' of a specific region corresponds to the `weights of the items' with respect to the concept {\it Fruits} in Table 1. Looking at Table 1, it is indeed {\it Apple}, which has the highest weight, equal to 0.1184, and hence is represented by spot number 8 on Figure 2, in the lightest region. Next comes {\it Elderberry} with weight equal to 0.1134, represented by spot number 7 on Figure 2, on the border of the lightest and second lightest region. Next comes {\it Raisin}, with weight equal to 0.1026, represented by spot number 6 on Figure 2, on the border of the third and the fourth lightest region. Next comes {\it Tomato}, with weight equal to 0.0881, represented by spot number 19 on Figure 2, in the seventh lightest region, etc. last is {\it Lentils}, with weight equal to 0.0095, represented by spot number 16 on Figure 2, in the one to darkest region. Hence Figure 2 contains 
a representation of the values of the collapse probability weights of the 24 items with respect to the concept {\it Fruits}. There is however more; we can, for example, wonder what the reason is to choose a representation in a plane? To explain this, turn to Figure 3. 

\begin{figure}[ht] 
\centering
\includegraphics[width=15cm]{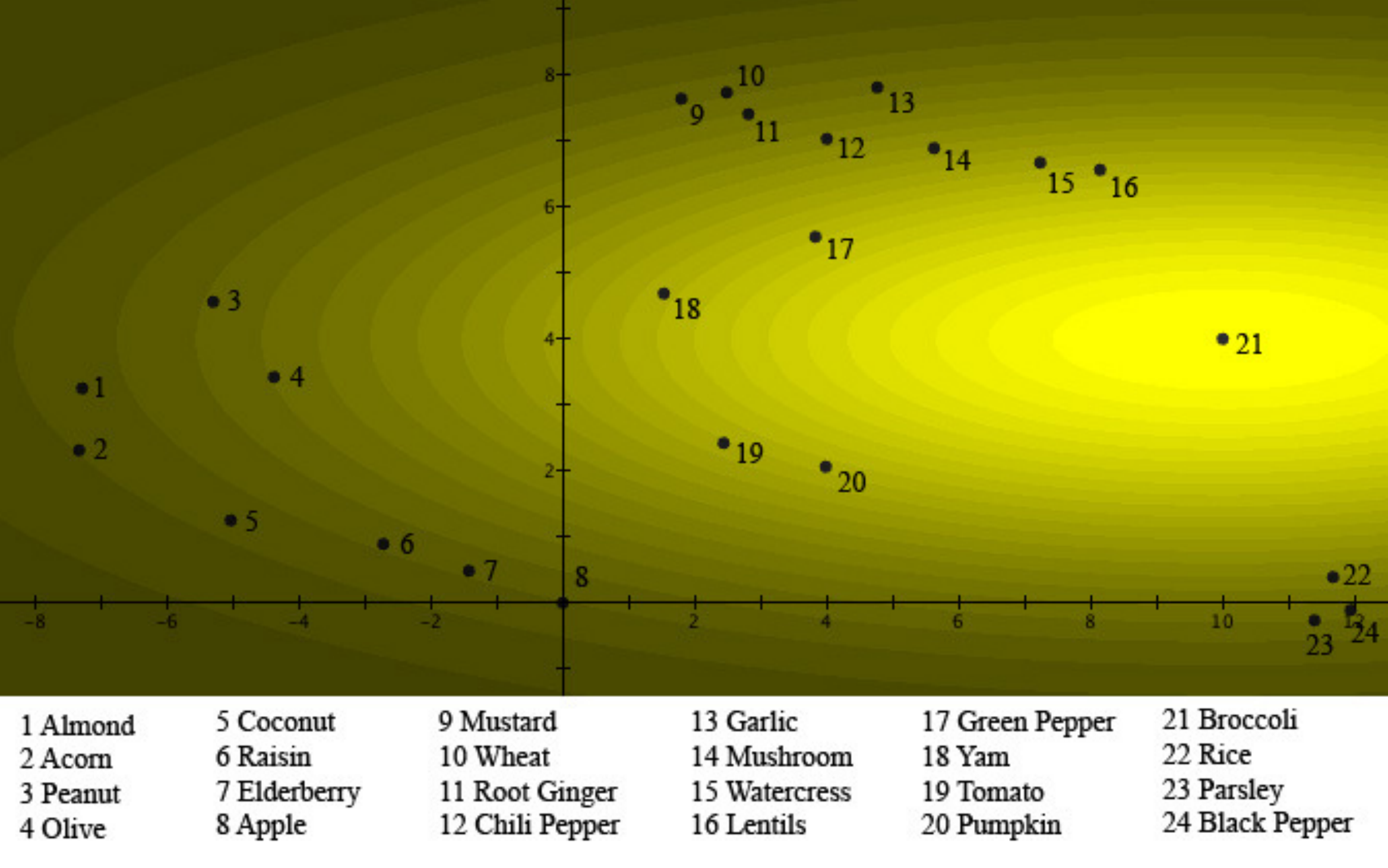}
\caption{The probabilities $\mu(B)_k$ of a person choosing the item $k$ as a `good example' of {\it Vegetables} are fitted into a two-dimensional quantum wave function $\psi_B(x, y)$. The numbers are placed at the locations of the different items with respect to the probability distribution $|\psi_B(x,y)|^2$. As in Figure 2, it can be seen as a light source shining through a hole centered on point 21, where {\it Broccoli} is located. The brightness of the light source in a specific region corresponds to the probability that this item will be chosen as a `good example' of {\it Vegetables}.} 
\end{figure}

Let us first note with respect to the two figures, although it might not seem the case at first sight, all the numbered regions are located at exactly the same spots in both Figure 2 and Figure 3, with respect to the two orthogonal axes that coordinate the plane. What is different in both figures are the graded structures of lighter to darker regions, while they are centered around the spot number 8, representing the item {\it Apple}, in Figure 2 they are centred around the spot number 21, representing the item {\it Broccoli}, in Figure 3. And, effectively, 
Figure 3 represents analogous to Figure 2 of the same 24 items, their collapse probability weights, but this time with respect to the concept {\it Vegetables}. This explains why in Figure 3 the lightest region is the one centered around spot number 21, representing {\it Broccoli}, while the lightest region in Figure 2 is the one centered around spot number 8, representing {\it Apple}. Indeed, {\it Broccoli} is the most characteristic vegetable of the considered items, while {\it Apple} is the most characteristic fruit, if `characteristic' is measured by the size of the respective collapse probability, i.e. the probability to choose this item in the course of the study. What might not seem obvious is that in a plane it is always possible to find 24 locations for the 24 items such that a graded structure with center {\it Apple} and a second graded structure with center {\it Broccoli} can be defined, fitting exactly also the other items in their correct value of `graded light to dark', corresponding to the collapse probability weights in Table 1. Such a situation is what we show in Figure 2 and in Figure 3. It can be proven mathematically that a solution always exists, although not a unique one, which means that Figures 2 and 3 show one of these solutions.

We have chosen on purpose the graded structure form light to dark to be colored yellow, because we can interpret Figures 2 and 3 such that an interesting analogy arises between our study of the 24 items and two concepts {\it Fruits} and {\it Vegetables}, and the well-known double slit experiment with light in quantum mechanics. It is this analogy that will also directly illustrate the `wave nature' of concepts. Suppose we consider a plane figuring in the experiment as a detection screen, and put counters for quantum light particles, i.e. photons, at the numbered spots on the plane. Then we send light through a first slit, which we call the {\it Fruits} slit, which is placed in front of the screen. The slit is placed such that the counters in the spots detect numbers of photons with fractions to the total number of photons send equal the collapse probability weights of the items represented by the respective spots with respect to the concept {\it Fruits}. The light received on the screen would then look like what is shown in Figure 2. Similarly, with counters placed in the same spots, we send light through a second slit, which we call the {\it Vegetable} slit. Now the counters detect numbers of photons with fractions to the total number of photons equal to the collapse probability weights of the same items with respect to the concept {\it Vegetables}. The light received on the screen would then look like what is shown in Figure 3. We can obtain the same figures directly for our psychological study, consisting of each participant choosing amongst the 24 items the one that he or she finds most characteristic of {\it Fruits} and {\it Vegetables} respectively. The relative frequencies of the first choice gives rise to the image in Figure 2, while the relative frequencies of the second choice gives rise to the image in Figure 3, if, for example, we would mark each chosen item by a fixed number of yellow light pixels on a computer screen. 

Before we combine the two slits to give rise to interference, let us specify the mathematics of the quantum mechanical formalism that underlies the two Figures. The situation can be represented quantum mechanically by complex valued Schr\"odinger wave functions of two real variables $\psi_A(x,y), \psi_B(x,y)$. For the light and the two slits, this situation is the `interaction of a photon with the two slits'. For the human participants in the concepts study, this situation is the `interaction with the two concepts of the mind of a participant'. We choose for $\psi_A(x,y)$ and $\psi_B(x,y)$ quantum wave packets, such that the radial part for both wave packets is a Gaussian in two dimensions. Considering Figures 2 and 3, we choose the top of the first Gaussian in the origin where spot number 8 is located, and the top of the second Gaussian in the point $(a,b)$, where spot number 21 is located. Hence
\begin{equation}
\psi_A(x,y)=\sqrt{D_A}e^{-({x^2 \over 4\sigma^2_{Ax}}+{y^2 \over 4\sigma^2_{Ay}})}e^{iS_A(x,y)} \qquad \psi_B(x,y)=\sqrt{D_B}e^{-({(x-a)^2 \over 4\sigma^2_{Bx}}+{(y-b)^2 \over 4\sigma^2_{By}})}e^{iS_B(x,y)}
\end{equation}
The phase parts of the wave packets $e^{iS_A(x,y)}$ and $e^{iS_B(x,y)}$ are determined by two phase fields $S_A(x,y)$ and $S_B(x,y)$ which will account for the interference and hence carry the wave nature. Of course, these phase parts vanish when we multiply each wave packet with its complex conjugate to find the connection with the collapse probabilities. Hence,
\begin{equation}
|\psi_A(x,y)|^2=D_Ae^{-({x^2 \over 2\sigma^2_{Ax}}+{y^2 \over 2\sigma^2_{Ay}})} \quad |\psi_B(x,y)|^2=D_Be^{-({(x-a)^2 \over 2\sigma^2_{Bx}}+{(y-b)^2 \over 2\sigma^2_{By}})}
\end{equation}
are the Gaussians to be seen in Figures 2 and 3, respectively.
Let us denote by $\Delta_k$ a small surface specifying the spot corresponding to the item number $k$ in the plane of the two figures. We then calculate the collapse probabilities of this item $k$ with respect to the concepts {\it Fruits} and {\it Vegetables} in a standard quantum mechanical way as follows
\begin{eqnarray} \label{firstintegral}
\mu_k(A)&=&\int_{\Delta_k}|\psi_A(x,y)|^2dxdy=\int_{\Delta_k}D_Ae^{-({x^2 \over 2\sigma^2_{Ax}}+{y^2 \over 2\sigma^2_{Ay}})}dxdy \\ \label{secondintegral}
\mu_k(B)&=&\int_{\Delta_k}|\psi_B(x,y)|^2dxdy=\int_{\Delta_k}D_Be^{-({x^2 \over 2\sigma^2_{Bx}}+{y^2 \over 2\sigma^2_{By}})}dxdy
\end{eqnarray}
We can prove that the parameters of the Gaussians, $D_A, \sigma_{Ax}, \sigma_{Ay}, D_B, \sigma_{Bx}, \sigma_{By}$ can be determined in such a way that the above equations come true, and for the images of Figures 2 and 3, exactly as we have done -- using an approximation for the integrals, which we explain later.

If we open both slits it will be the normalized superposition of the two wave packets that quantum mechanically describes the new situation
\begin{equation}
\psi_{A{\rm or}B}(x,y)={1 \over \sqrt{2}}(\psi_A(x,y)+\psi_B(x,y))
\end{equation}
We have
\begin{eqnarray}
\mu_k(A\ {\rm or}\ B)&=&\int_{\Delta_k}\psi_{A{\rm or}B}(x,y)^*\psi_{A{\rm or}B}(x,y)dxdy \nonumber \\
&=&{1 \over 2}(\int_{\Delta_k}\psi_A(x,y)^*\psi_A(x,y)dxdy+\int_{\Delta_k}\psi_B(x,y)^*\psi_B(x,y)dxdy) \nonumber \\
&&+\int_{\Delta_k}\Re(\psi_A(x,y)^*\psi_B(x,y))dxdy \nonumber \\
&=&{1 \over 2}(\mu_k(A)+\mu_k(B))+\int_{\Delta_k}\Re(\psi_A(x,y)^*\psi_B(x,y))dxdy \label{integralequation01}
\end{eqnarray}
Let us calculate $\int_{\Delta_k}\Re(\psi_A(x,y)^*\psi_B(x,y))dxdy$. We have
\begin{eqnarray}
&&\int_{\Delta_k}\Re(\psi_A(x,y)^*\psi_B(x,y))dxdy \nonumber \\
&&=\int_{\Delta_k}(\sqrt{D_A}e^{-({x^2 \over 4\sigma^2_{Ax}}+{y^2 \over 4\sigma^2_{Ay}})})(\sqrt{D_B}e^{-({(x-a)^2 \over 4\sigma^2_{Bx}}+{(y-b)^2 \over 4\sigma^2_{By}})})\Re(e^{-iS_A(x,y)}e^{iS_B(x,y)})dxdy \nonumber \\
&&=\int_{\Delta_k}(\sqrt{D_AD_B}e^{-({x^2 \over 4\sigma^2_{Ax}}+{(x-a)^2 \over 4\sigma^2_{Bx}}+{y^2 \over 4\sigma^2_{Ay}}+{(y-b)^2 \over 4\sigma^2_{By}})})\Re(e^{i(S_B(x,y)-S_A(x,y))})dxdy \nonumber \\
&&=\int_{\Delta_k}(\sqrt{D_AD_B}e^{-({x^2 \over 4\sigma^2_{Ax}}+{(x-a)^2 \over 4\sigma^2_{Bx}}+{y^2 \over 4\sigma^2_{Ay}}+{(y-b)^2 \over 4\sigma^2_{By}})})\cos(S_B(x,y)-S_A(x,y))dxdy
\end{eqnarray}
We can hence rewrite equation (\ref{integralequation01}) in the following way
\begin{eqnarray}
\int_{\Delta_k}f(x,y)\cos\theta(x,y)dxdy=f_k \label{integralequation02}
\end{eqnarray}
where
\begin{equation}
f(x,y)=\sqrt{D_AD_B}e^{-({x^2 \over 4\sigma^2_{Ax}}+{(x-a)^2 \over 4\sigma^2_{Bx}}+{y^2 \over 4\sigma^2_{Ay}}+{(y-b)^2 \over 4\sigma^2_{By}})}
\end{equation}
is a known Gaussian-like function, remember that we have determined $D_A$, $D_B$, $\sigma_{Ax}$, $\sigma_{Ay}$, $\sigma_{Bx}$, $\sigma_{By}$ and $a$ and $b$ in choosing a solution to be seen in 
Figures 2 and 3, and 
\begin{equation}
f_k=\mu_k(A\ {\rm or}\ B)-{1 \over 2}(\mu_k(A)+\mu_k(B))
\end{equation}
are constants for each $k$ determined by the data, and we have introduced
\begin{equation}
\theta(x,y)=S_B(x,y)-S_A(x,y)
\end{equation}
the field of phase differences of the two quantum wave packets. This field of phases differences will determine the interference pattern and it is the solution of the 24 nonlinear equations in (\ref{integralequation02}). This set of 24 equations cannot be solved exactly, but even a general numerical solution is not straightforwardly at reach within actual optimization programs. We have introduces two steps of idealization to find a solution. First, we have looked for a solution where $\theta(x,y)$ is a large enough, polynomial in $x$ and $y$, more specifically consisting of 24 independent sub-polynomials that are independent
\begin{eqnarray}
\theta(x,y)&=&F_1+F_2x+F_3y+F_4x^2+F_5xy+F_6y^2+F_7x^3+F_8x^2y+F_9xy^2+F_{10}y^3 \nonumber \\
&&+F_{11}x^4+F_{12}x^3y+F_{13}x^2y^2+F_{14}xy^3+F_{15}y^4+F_{16}x^5+F_{17}x^4y+F_{18}x^3y^2 \nonumber \\
&&+F_{19}x^2y^3+F_{20}xy^4+F_{21}y^5+F_{22}x^6+F_{23}x^5y+F_{24}x^4y^2 \label{phasefield}
\end{eqnarray}
Secondly, we suppose that $\Delta_k=\Delta$ is a sufficiently small square surface such that a good approximation of the integral in (\ref{integralequation02}) -- and it is also the approximation we have used for the integrals (\ref{firstintegral}) and (\ref{secondintegral}) -- is given by $\Delta$ times the value of the function under the integral in the centre of $\Delta$. This transforms the set of 24 nonlinear equations (\ref{integralequation02}) into a set of 24 linear equations
\begin{equation}
\Delta f(x_k,y_k)\theta(x_k,y_k)=f_k
\end{equation}
We have solved them for the points $(x_k,y_k)$ where the 24 items are located in Figures 2 and 3, for $\Delta=0.01$, which gives us $\theta(x,y)$, and hence also the expression for $|\psi_{A{\rm or}B}(x,y)|^2$ containing the expected interference term, and we have
\begin{equation}
|\psi_{A{\rm or}B}(x,y)|^2={1 \over 2}(|\psi_A(x,y)|^2+|\psi_B(x,y)|^2)+|\psi_A(x,y)\psi_B(x,y)|\cos\theta(x,y)
\end{equation}

\begin{figure}[ht] 
\centering
\includegraphics[width=15cm]{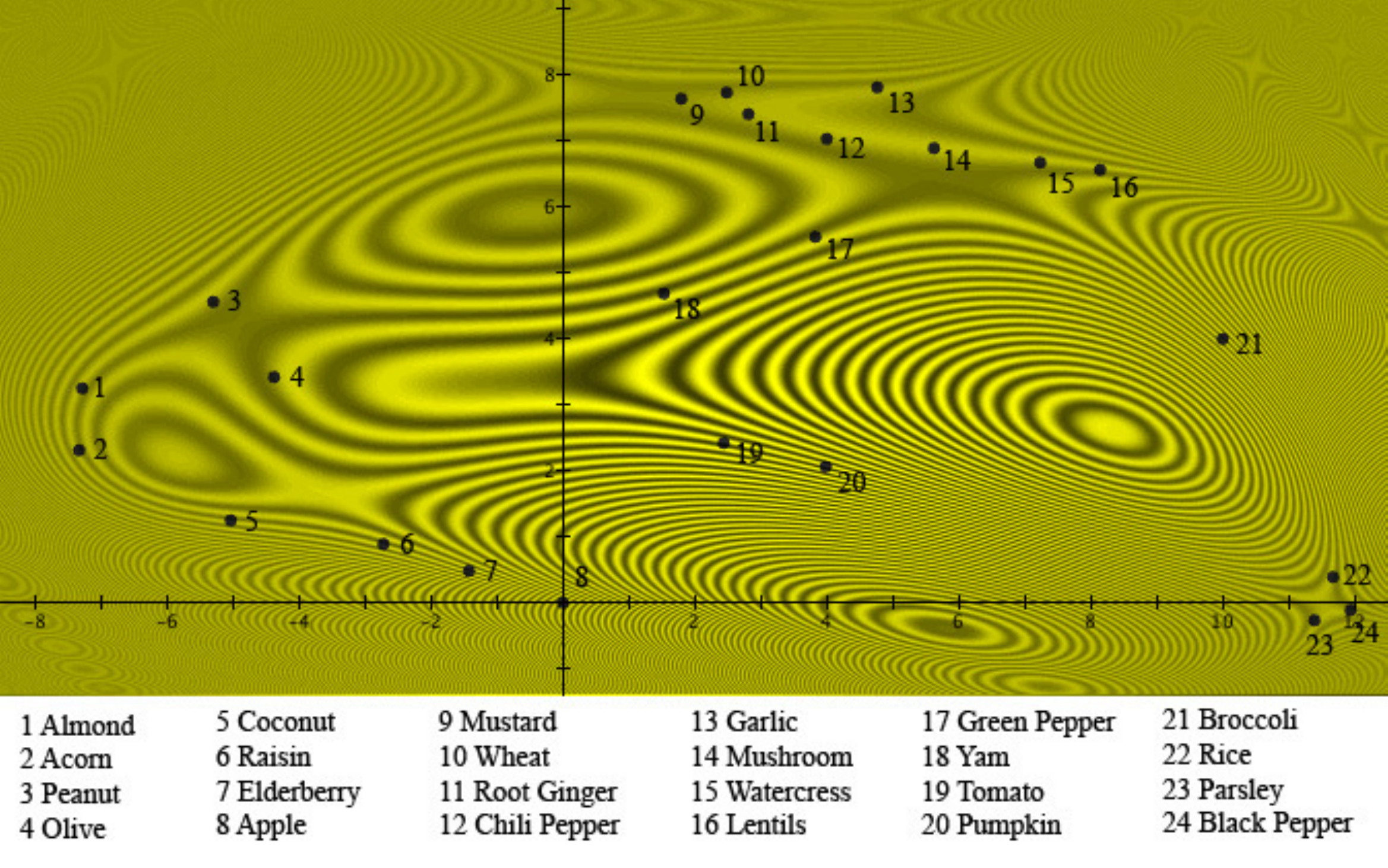}
\caption{The probabilities $\mu(A\ or\ B)_k$ of a person choosing the item $k$ as a `good example' of {\it Fruits or Vegetables} are fitted into the two-dimensional quantum wave function ${1 \over \sqrt{2}}(\psi_A(x,y)+\psi_B(x,y))$, which is the normalized superposition of the wave functions in Figures 2 and 3. The numbers are placed at the locations of the different exemplars with respect to the probability distribution $|\psi_A(x,y)+\psi_B(x,y)|^2 = {1 \over 2}(|\psi_A(x,y)|^2 +|\psi_B(x,y)|^2)+|\psi_A(x,y)\psi_B(x,y)|\cos\theta(x,y)$, where $\theta(x,y)$ is the quantum phase difference at $(x, y)$. The values of $\theta(x, y)$ are given in Table 1 for the locations of the different items. The interference pattern is clearly visible.} 
\end{figure}

In Figure 4 we have graphically represented this probability density $|\psi_{A{\rm or}B}(x,y)|^2$. The interference pattern shown in Figure 4 is very similar to well-known interference patterns of light passing through an elastic material under stress. In our case, it is the interference pattern corresponding to `Fruits or Vegetables' as a contextual, interfering prototype. The numerical values of the solutions represented in Figures 2, 3 and 4 are in Table 2.

 \begin{table}[ht]
  \begin{center}
  \begin{tabular}{@{} clccccrrr @{}}
    \hline
               & &  Parameters of the solution                    &                                           &           & &    &         \\
    \hline
   $k$  &  item & $(x,y)$ coordinates of items  & sub-polynomial & coefficients $F_k$ & Gaussian & parameters  \\ 
    \hline
    1 & Almond & (-7.2826, 3.24347) & $1$ & 87.6039 & $D_A$ & 1.18412 \\ 
    2 & Acorn & (-7.3316, 2.3116) & $x$ &  2792.02 & $\sigma_{A_x}$ & 5.65390  \\ 
    3 & Peanut & (-5.2957, 4.56032) & $y$ & 8425.01 & $\sigma_{A_y}$ & 3.80360   \\ 
    4 & Olive & (-4.3776, 3.41765) & $x^2$ & 19.36 & $D_B$ & 1.28421  \\ 
    5 & Coconut & (-5.0322, 1.24573) & $xy$ & -2139.87  & $\sigma_{B_x}$ & 8.20823  \\ 
    6 & Raisin & (-2.7149, 0.896651) & $y^2$ & -7322.26  & $\sigma_{B_y}$ & 2.41578  \\ 
    7 & Elderberry & (-1.420, 0.487598) & $x^3$ & -39.2811 & &  \\ 
    8 & Apple & (0, 0) & $x^2y$ & -55.5263 & &    \\ 
    9 & Mustard & (1.7978, 7.64549) & $xy^2$ & 586.674 & &  \\ 
   10 & Wheat & (2.4786, 7.73915) & $y^3$ & 2205.81 & &   \\ 
   11 & Root Ginger & (2.8164, 7.41004) & $x^4$ & -2.22868 & &  \\ 
    12 & Chili Pepper & (3.9933, 7.03549) & $x^3y$ & 4.19408 & &  \\ 
   13 & Garlic & (4.7681, 7.81803) & $x^2y^2$ & 13.3579 & & \\ 
    14& Mushroom & (5.6281, 6.89107) & $xy^3$ & -72.233 & &   \\ 
   15 & Watercress & (7.233, 6.67322) & $y^4$ & -275.834  & &  \\ 
   16 & Lentils & (8.1373, 6.56281) & $x^5$ & 0.426731 & &  \\ 
    17 & Green Pepper & (3.8337, 5.55379) & $x^4y$ & 1.58764 & & \\ 
   18 & Yam & (1.5305, 4.69497) & $x^3y^2$ & 0.582536 & & \\ 
   19 & Tomato & (2.4348, 2.42612) & $x^2y^3$ & -1.13167 & & \\ 
    20 & Pumpkin & (3.9873, 2.06652) & $xy^4$ & 3.44008 & &  \\ 
    21 & Broccoli & (10, 4) & $y^5$ & 12.2584 & & \\ 
    22 & Rice & (11.6771, 0.392458) & $x^6$ & -0.00943132 & & \\ 
   23 & Parsley & (11.3949, -0.268463) & $x^5y$ & -0.0535881 & &  \\ 
   24 & Black Pepper & (11.9389, -0.107151) & $x^4y^2$ & -0.200688 & &  \\ 
    \hline
  \end{tabular}
  \end{center}

\vspace{1mm}
\noindent
 {\bf Table 2.} 
The parameters of the interference pattern solution illustrated in Figure 4. The first column lists the different items, and the second column the coordinates of their locations in Figure 2, Figure 3 and Figure 4. The third column contains the orthogonal set of sub-polynomials used as first approximation for the phase field $\theta(x,y)$, and the fourth column their values. The fifth and sixth columns contain the Gaussian parameters and their values of the solution.
\end{table}

We have thus completed our illustration of contextual interfering prototypes. It is, however, important to remember 
that this representation is at the subtle level of an 
illustration, while the real working representation of contextual interfering prototypes needs the complete quantum-mechanical formalism. It can be considered as a pre-representation, exactly as the wave-like representations by de Broglie and Schr\"{o}dinger in the early days of quantum physics can be considered as useful pre-quantum representations that capture something of the wave aspects of microscopic particles.

\section{Discussion}\label{conclusions}
In this paper we showed that a generalization of prototype theory can address the `Pet-Fish problem' and related combination issues. This was done by formalizing the effect of the cognitive context on the state of a concept using a SCoP formalism 
\cite{ga2002,ag2005a,ag2005b,gra2008}. We also developed a quantum-theoretic model in complex Hilbert space to show that, in this contextualized prototype theory, prototypes can interfere when 
concepts combine, as evidenced by data where typicality measurements are performed. This could then lead one to think that the general quantum approach to concepts only presupposes a (contextual) prototype theory. 
We now explain why 
this inference is not true. 

Let us make more explicit the relationship between our quantum-conceptual approach and other 
concept theories,
such as prototype theory, exemplar theory and theory theory. A deeper analysis shows that our approach is more than a contextual generalization of prototype theory. Roughly speaking, other theories make assumptions 
about the principles guiding the formation and intuitive representation of a concept in the human mind. Thus, prototype theory assumes that a concept is
determined by a set of characteristic rather than defining features, the human mind has a privileged prototype for each concept, and typicality of a concrete item is determined by its similarity with the prototype \cite{r1973,r1978,r1983}. Exemplar theory assumes instead that a concept is not determined by a set of defining or characteristic features but, rather, by a set of salient instances of it stored in memory \cite{n1988,n1992}. Theory theory assumes that 
concepts are determined by `mini-theories' or schemata, identifying the causal relationships among properties \cite{mm1985,rn1988}. 
These theories have all mainly been preoccupied with the question of `what predominantly determines a concept'. We agree on the relevance of this question, though it is not the main issue focused on there. Transposed to our approach, these theories mainly investigate `what predominantly determines the state of a concept'. Conversely, the main preoccupation of our approach has been to propose a theory with the following features:

(i) a well-defined ontology, i.e. a concept is in our approach an entity capable of different modes of being with respect to how it influences measurable semantic quantities such as typicality, membership weight and membership probability, and these modes are called `states';

(ii) the capacity to produce theoretical models fitting data on these measurable semantic quantities. 

We seek to achieve (i) and (ii) independent of the question that is the focus of 
other theories of concepts. More concretely, and in accordance with the results of investigations into the question of `what predominantly determines a concept', as far as prototype theory, exemplar theory and theory theory are concerned, we believe that all approaches are partially valid. The state of a concept, i.e. its capability of influencing the values of measurable semantic quantities, such as typicality and membership weight, is influenced by the set of its characteristic features, but also by salient exemplars in memory, and in a considerable number of cases -- where more causal aspects are at play -- mini-theories might be appropriate to express this state. It is important that `a conceptual state is defined and gives rise, together with the context, to the values of the measurable semantic quantities'. The fact that the specification of these values can be only probabilistic is a confirmation that potentiality and uncertainty occur even if the state is completely known, hence quantum structures are intrinsically needed.

It follows from the above that resorting and giving new life to prototype theory does not necessarily entail that contextual prototype theory is the only possible theory of concepts for what concerns the question of `what predominantly determines a concept'. However, we choose to identify our general approach as a `generalized contextual interfering prototype theory', because the `ground state' of a concept is a fundamental notion of the theory, and this ground state is what corresponds to the prototype. There is not a similar affinity with exemplar theory and theory theory. However, the conceptual state and its interaction with the cognitive context can potentially capture the other conceptual aspects, exemplars and schemata, which are instead predominant in alternative concept theories. In this respect, an interesting analogy must be emphasized. The quantum-theoretic approach only aims at modeling concepts and their combinations in a unitary and coherent mathematical formalism. We do not pretend to give a universal definition of what a concept is and how it forms. Using a known analogy in mathematics, we can say that the quantum-theoretic model is to a concept as a traditional Kolmogorov model is to a probability. A Kolmogorovian model specifies how a probability can be mathematically formalized independent of the definition of probability that is chosen (favorable over possible cases, large number limit of frequencies, subjective, etc.). Analogously, the quantum-theoretic framework for concepts enables mathematical modeling of conceptual entities independent of the definition that is adopted in a specific concept theory (prototype, exemplar, theory, etc.).

We conclude with an epistemological consideration. The quantum-theoretic framework presented here constitutes a step toward the elaboration of a general theory for the representation of any conceptual entity. Hence, it is not just a `cognitive model for 
typicality, membership weight or membership probability'. Rather, we are investigating whether `quantum theory, in its Hilbert space formulation, is an appropriate theory to model human cognition'. To understand what we mean by this let us consider an example taken from everyday life. As an example of a theory, we could introduce the theory of `how to make good clothes'. A tailor needs to learn how to make good clothes for different types of people, men, women, children, old people, etc. Each cloth is a model in itself. Then, one can also consider intermediate situations where one has models of series of clothes. A specific body will not fit in any clothes: you need to adjust the parameters (length, size, etc.) to reach the desired fit. We think that  a theory should be able to reproduce different experimental results by suitably adjusting the involved parameters, exactly as a theory of clothing. This is 
different from a set of models, even if the set can cope with a wide range of data.

There is a tendency, mainly in empirically-based disciplines, to be critical with respect to a theory that can cope with all possible situations it applies to. This is because the theory contains too many parameters, which may lead one to think that `any type of data can be modeled by allowing all these parameters to have different values'. We agree that, in case we have to do with an `ad hoc model', i.e. a model specially made for the circumstance of the situation it models, this suspicion is grounded. Adding parameters to such an ad hoc model, or stretching the already contained parameters to other values, does not give rise to what we call a theory. On the other hand, a theory needs to be well defined, its rules, the allowed procedures, its theoretical, mathematical, and internal logical structure, `independent' of the structure of the models describing specific situations that can be coped with by the theory. Hence also the theory needs to contain a well defined description of `how to produce models for specific situations'. Coming back to the theory of clothing, if a tailor knows the theory of clothing, obviously he or she can make a cloth for every human body, because the theory of clothing, although its structure is defined independently of a specific cloth, contains a prescription of how to apply it to any possible specific cloth. In this respect, we think that one should carefully distinguish between a model that is derived by a general theory, as the one presented in this paper, and a model specifically designed to test a number of experimental situations.

This brings us to the important question of the `predictive power' of existing quantum-theoretic models. Models derived from a theory will generally need more data from a bigger set of experiments to become predictive for the outcomes of other not yet performed experiments than this is the case for models that are more ad hoc. The reason is that in principle such models -- think of the analogy we present with the theory of clothing above -- must be able to faithfully represent the data of all possible experiments that can be performed on the conceptual entity in the same state. A tailor knowing the theory of clothing can in principle make clothes for all human bodies but hence also predicts outcomes of not performed experiments, e.g., the measure of a specific part of the cloth, if enough data of a set of experiments are available to the tailor, e.g., data that determine the possible types of clothes still fitting these data and as a consequence also determine the measure of this part of the clothe. In general in quantum cognition, the scarcity of data is preventing models from having systematic and substantial predictive power. One can wonder, if predictive power is not yet predominantly available in the majority of existing quantum-theoretic models, why so much attention and value is actually attributed to them? Answering this question allows us to clarify an aspect of quantum cognition that is not obvious and even makes it special in a specific way, at least provisionally until more data is available. 
The success of quantum cognition is due to it `being able to convincingly model data that theoretically can be proven to be impossible to model with any model that relies on classical fuzzy set theory and/or classical Kolmogorovian probability theory'. Hence, a different criterion than predictive power is provisionally used to identify the success of quantum cognition. Of course, as soon as more data are collected, the models will also be able to be tested for their predictive power. Recent work in quantum cognition is starting to reach the level of being predictive, for example study of order effects \cite{wangetal2014}, and an elaboration and refinement of the model presented in this article\cite{asv2015,PhilTransA2015}. The latter model simultaneously investigates the `conjuntion' and the `negation' of concepts, starting from data collected on such conceptual combinations. To explain the exact nature and also accurateness of the predictive power we gained in the model in \cite{asv2015,PhilTransA2015}, consider the following mathematical expression
\begin{equation}
I_{ABA'B'}= 1-\mu(A\ {\rm and}\ B)-\mu(A\ {\rm and}\ B')-\mu(A'\ {\rm and}\ B)-\mu(A'\ {\rm and}\ B') \label{classicalexpression}
\end{equation}  
where $A$ and $B$ are the concepts {\it Fruits} and {\it Vegetables}, respectively, while $A'$ and $B'$ are their negations. Thus, `$A$ and $B'$' means {\it Fruits and not Vegetables}, while `$A'$ and $B$' means {\it not Fruits and Vegetables} and `$A'$ and $B'$' means {\it not Fruits and not Vegetables}. In \cite{asv2015,PhilTransA2015} we published the data for the outcomes of experiments that test the membership of the same 24 items which we considered in the present article, but this time not only for the conjunction of $A$ and $B$, but also for the conjunctions `$A$ and $B'$', `$A'$ and $B$' and `$A'$ and $B'$'. Suppose that the data follow a classical probabilistic structure, then $I_{ABA'B'}$ has to be theoretically equal to zero for each considered item, and this purely follows from a general `law of probability calculus' related to the so called `de Morgan laws' of classical probability. This means that, under the hypothesis of a classical probabilistic structure, if we measure the relative frequencies of `$A$ and $B$', `$A$ and $B'$' and `$A'$ and $B$', and hence determine experimentally the values of $\mu(A\ {\rm and}\ B)$, $\mu(A\ {\rm and}\ B')$ and $\mu(A'\ {\rm and}\ B)$, a `prediction' for $\mu(A'\ {\rm and}\ B')$ can be made theoretically, namely,
\begin{equation}
\mu(A'\ {\rm and}\ B')=1-\mu(A\ {\rm and}\ B)-\mu(A\ {\rm and}\ B')-\mu(A'\ {\rm and}\ B) \label{predictionclassical}
\end{equation} 
for each considered item.  
Let is explain what are our findings in \cite{asv2015,PhilTransA2015} that make it possible for us to speak of some specific type of predictability for the more elaborated and refined model we developed for the combination of concepts and their negations.
In \cite{asv2015,PhilTransA2015} we have collected data not only for the pair of concepts {\it Fruits} and {\it Vegetables} and the 24 items treated also in the present article, but for three more pairs of concepts, and for each of them again 24 items. Due to the already identified non classical nature of overextension of the conjunction we expected that $I_{ABA'B'}$ would not be equal to zero, and that indeed showed to be the case. However, we detected a high level of systematics of the value of $I_{ABA'B'}$ fluctuating around an average of $-0.81$. A statistical analysis showed the different values for individual items to be possible to be explained as fluctuations around this average (see Tables 1, 2, 3 and 4 in \cite{asv2015}). Next to the detailed statistical analysis to be found in \cite{asv2015} we also put forward a theoretical explanation of this value. The elaborated and refined model for concept combinations developed in\cite{asv2015} introduces within the model the combination of a pure quantum model and a classical model. It can be shown that for a pure quantum model the value of $I_{ABA'B'}$ would be $-1$. We also find that the quantum effects are dominant as compared to the classical effects in case concepts are combined, which explains why our refined model gives rise to a value of $I_{ABA'B'}$ in between the classical one, which is $0$, and the pure quantum one, which is $-1$, but closer to the quantum one, hence $-0.81$. This finding can be turned into a predictive one in the following way. Suppose we measure $\mu(A\ {\rm and}\ B)$, $\mu(A\ {\rm and}\ B')$ and $\mu(A'\ {\rm and}\ B)$ for two arbitrary concepts and an item. Our model allows us to put forward the following prediction for $\mu(A'\ {\rm and}\ B')$
\begin{equation} \label{predictionquantum}
\mu(A'\ {\rm and}\ B')=1.81-\mu(A\ {\rm and}\ B)-\mu(A\ {\rm and}\ B')-\mu(A'\ {\rm and}\ B)
\end{equation}
By comparing (\ref{predictionclassical}) and (\ref{predictionquantum}), we get that the quantum-theoretic model in \cite{asv2015} provides a `different prediction' from a classical probabilistic model satisfying the axioms of Kolmogorov, and experiments confirm the validity of the former over the latter. We add that the quantum model has different predictions from a classical model also for the values of other functions than $I_{ABA'B'}$, and these predictions are `parameter independent', in the sense that they do not depend on the values of free parameters that may accommodate the data.

The results above can be considered as a strong confirmation that quantum-theoretic models of concept combinations provide predictions that deviate, in some situations, from the predictions of classical Kolmogorovian models, which is confirmed by experimental data.


\appendix

\section*{Appendix. Quantum mathematics for conceptual modeling}\label{quantum}
We illustrate in this section how the mathematical formalism of quantum theory can be applied to model situations outside the microscopic quantum world, more specifically, in the representation of concepts and their combinations. We will limit technicalities to the essential.

When the quantum mechanical formalism is applied for modeling purposes, each considered entity  -- in our case a concept -- is associated with a complex Hilbert space ${\mathcal H}$, that is, a vector space over the field ${\mathbb C}$ of complex numbers, equipped with an inner product $\langle \cdot |  \cdot \rangle$ that maps two vectors $\langle A|$ and $|B\rangle$ onto a complex number $\langle A|B\rangle$. We denote vectors by using the bra-ket notation introduced by Paul Adrien Dirac, one of the pioneers of quantum theory. Vectors can be `kets', denoted by $\left| A \right\rangle $, $\left| B \right\rangle$, or `bras', denoted by $\left\langle A \right|$, $\left\langle B \right|$. The inner product between the ket vectors $|A\rangle$ and $|B\rangle$, or the bra-vectors $\langle A|$ and $\langle B|$, is realized by juxtaposing the bra vector $\langle A|$ and the ket vector $|B\rangle$, and $\langle A|B\rangle$ is also called a `bra-ket', and it satisfies the following properties:

(i) $\langle A |  A \rangle \ge 0$;

(ii) $\langle A |  B \rangle=\langle B |  A \rangle^{*}$, where $\langle B |  A \rangle^{*}$ is the complex conjugate of $\langle A |  B \rangle$;

(iii) $\langle A |(z|B\rangle+t|C\rangle)=z\langle A |  B \rangle+t \langle A |  C \rangle $, for $z, t \in {\mathbb C}$,
where the sum vector $z|B\rangle+t|C\rangle$ is called a `superposition' of vectors $|B\rangle$ and $|C\rangle$ in the quantum jargon.

From (ii) and (iii) follows that inner product $\langle \cdot |  \cdot \rangle$ is linear in the ket and anti-linear in the bra, i.e. $(z\langle A|+t\langle B|)|C\rangle=z^{*}\langle A | C\rangle+t^{*}\langle B|C \rangle$.

The `absolute value' of a complex number is defined as the square root of the product of this complex number times its complex conjugate, that is, $|z|=\sqrt{z^{*}z}$. Moreover, a complex number $z$ can either be decomposed into its cartesian form $z=x+iy$, or into its polar form $z=|z|e^{i\theta}=|z|(\cos\theta+i\sin\theta)$. As a consequence, we have $|\langle A| B\rangle|=\sqrt{\langle A|B\rangle\langle B|A\rangle}$. We define the `length' of a ket (bra) vector $|A\rangle$ ($\langle A|$) as $|| |A\rangle ||=||\langle A |||=\sqrt{\langle A |A\rangle}$. A vector of unitary length is called a `unit vector'. We say that the ket vectors $|A\rangle$ and $|B\rangle$ are `orthogonal' and write $|A\rangle \perp |B\rangle$ if $\langle A|B\rangle=0$.

We have now introduced the necessary mathematics to state the first modeling rule of quantum theory, as follows.

\medskip
\noindent{\it First quantum modeling rule:} A state $A$ of an entity -- in our case a concept -- modeled by quantum theory is represented by a ket vector $|A\rangle$ with length 1, that is $\langle A|A\rangle=1$.

\medskip
\noindent
An orthogonal projection $M$ is a linear operator on the Hilbert space, that is, a mapping $M: {\mathcal H} \rightarrow {\mathcal H}, |A\rangle \mapsto M|A\rangle$ which is Hermitian and idempotent. The latter means that, for every $|A\rangle, |B\rangle \in {\mathcal H}$ and $z, t \in {\mathbb C}$, we have:

(i) $M(z|A\rangle+t|B\rangle)=zM|A\rangle+tM|B\rangle$ (linearity);

(ii) $\langle A|M|B\rangle=\langle B|M|A\rangle^{*}$ (hermiticity);

(iii) $M \cdot M=M$ (idempotency).

The identity operator $\mathbbmss{1}$ maps each vector onto itself and is a trivial orthogonal projection. We say that two orthogonal projections $M_k$ and $M_l$ are orthogonal operators if each vector contained in $M_k({\mathcal H})$ is orthogonal to each vector contained in $M_l({\mathcal H})$, and we write $M_k \perp M_l$, in this case. The orthogonality of the projection operators $M_{k}$ and $M_{l}$ can also be expressed by $M_{k}M_{l}=0$, where $0$ is the null operator. A set of orthogonal projection operators $\{M_k\ \vert k=1,\ldots,n\}$ is called a `spectral family' if all projectors are mutually orthogonal, that is, $M_k \perp M_l$ for $k \not= l$, and their sum is the identity, that is, $\sum_{k=1}^nM_k=\mathbbmss{1}$.

The above definitions give us the necessary mathematics to state the second modeling rule of quantum theory, as follows.

\medskip
\noindent
{\it Second quantum modeling rule:} A measurable quantity $Q$ of an entity -- in our case a concept -- modeled by quantum theory, and having a set of possible real values $\{q_1, \ldots, q_n\}$ is represented by a spectral family $\{M_k\ \vert k=1, \ldots, n\}$ in the following way. If the entity -- in our case a concept -- is in a state represented by the vector $|A\rangle$, then the probability of obtaining the value $q_k$ in a measurement of the measurable quantity $Q$ is $\langle A|M_k|A\rangle=||M_k |A\rangle||^{2}$. This formula is called the `Born rule' in the quantum jargon. Moreover, if the value $q_k$ is actually obtained in the measurement, then the initial state is changed into a state represented by the vector
\begin{equation}
|A_k\rangle=\frac{M_k|A\rangle}{||M_k|A\rangle||}
\end{equation}
This change of state is called `collapse' in the quantum jargon.

\medskip
\noindent
The tensor product ${\mathcal H}_{A} \otimes {\mathcal H}_{B}$ of two Hilbert spaces ${\mathcal H}_{A}$ and ${\mathcal H}_{B}$ is the Hilbert space generated by the set $\{|A_i\rangle \otimes |B_j\rangle\}$, where $|A_i\rangle$ and $|B_j\rangle$ are vectors of ${\mathcal H}_{A}$ and ${\mathcal H}_{B}$, respectively, which means that a general vector of this tensor product is of the form $\sum_{ij}|A_i\rangle \otimes |B_j\rangle$. This gives us the necessary mathematics to introduce the third modeling rule.

\medskip
\noindent
{\it Third quantum modeling rule:} A state $C$ of a compound entity -- in our case a combined concept -- is represented by a unit vector $|C\rangle$ of the tensor product ${\mathcal H}_{A} \otimes {\mathcal H}_{B}$ of the two Hilbert spaces ${\mathcal H}_{A}$ and ${\mathcal H}_{B}$ containing the vectors that represent the states of the component entities -- concepts.

\medskip
\noindent
The above means that we have $|C\rangle=\sum_{ij}c_{ij}|A_i\rangle \otimes |B_j\rangle$, where $|A_i\rangle$ and $|B_j\rangle$ are unit vectors of ${\mathcal H}_{A}$ and ${\mathcal H}_{B}$, respectively, and $\sum_{i,j}|c_{ij}|^{2}=1$. We say that the state $C$ represented by $|C\rangle$ is a product state if it is of the form $|A\rangle \otimes |B\rangle$ for some $|A\rangle \in {\mathcal H}_{A}$ and $|B\rangle \in {\mathcal H}_{B}$. Otherwise, $C$ is called an `entangled state'.

\medskip
\noindent
The Fock space is a specific type of Hilbert space, originally introduced in quantum field theory. For most states of a quantum field the number of identical quantum entities is not conserved but is a variable quantity. The Fock space copes with this situation in allowing its vectors to be superpositions of vectors pertaining to different sectors for fixed numbers of identical quantum entities. MoreA explicitly, the $k$-th sector of a Aock space describes a fixed number of $k$ identical quantum entities, and it is of the form ${\mathcal H}\otimes \ldots \otimes{\mathcal H}$ of the tensor product of $k$ identical Hilbert spaces ${\mathcal H}$. The Aock space $A$ itself is the direct sum of all these sectors, hence
\begin{equation} \label{fockspace}
{\mathcal A}=\oplus_{k=1}^j\otimes_{l=1}^k{\mathcal H}
\end{equation}
Aor our modeling we have only used Aock space for the `two' and `one quantum entity' case, hence ${\mathcal A}={\mathcal H}\oplus({\mathcal H}\otimes{\mathcal H})$. This is due to considering only combinations of two concepts. The sector ${\mathcal H}$ is called the `first sector', while the sector ${\mathcal H}\otimes{\mathcal H}$ is called the `second sector'. A unit vector $|F\rangle \in {\mathcal F}$ is then written as $|F\rangle = ne^{i\gamma}|C\rangle+me^{i\delta}(|A\rangle\otimes|B\rangle)$, where $|A\rangle, |B\rangle$ and $|C\rangle$ are unit vectors of ${\mathcal H}$, and such that $n^2+m^2=1$. For combinations of $j$ concepts, the general form of Fock space  in  (\ref{fockspace}) should be used.

The quantum modeling above can be generalized by allowing states to be represented by the so called `density operators' and measurements to be represented by the so called `positive operator valued measures'. However, for the sake of brevity we will not dwell on this extension here.


\end{document}